\documentclass[preprint,12pt]{elsarticle}

\usepackage{amssymb}
\usepackage{amsmath}
\usepackage{subfig}
\usepackage{multirow}
\usepackage{booktabs}
\usepackage{siunitx}
\usepackage[table]{xcolor}
\usepackage{url} 
\usepackage{booktabs}
\usepackage{makecell}

\begin{document}

\begin{frontmatter}


\title{Domain-Adapted Pre-trained Language Models for Implicit Information Extraction in Crash Narratives}

\author[inst1]{Xixi Wang\corref{cor1}}
\ead{hixixi66@gmail.com}
\cortext[cor1]{Corresponding author}

\author[inst2]{Jordanka Kovaceva}


\author[inst1]{Miguel Costa}
\author[inst3]{Shuai Wang}

\author[inst1]{Francisco Camara Pereira}
\author[inst2]{Robert Thomson}

\affiliation[inst1]{organization={Department of Technology, Management and Economics, Technical University of Denmark},
            addressline={Akademivej}, 
            city={2800 Kongens Lyngby},
            postcode={}, 
            country={Denmark}}

\affiliation[inst2]{organization={Department of Mechanics and Maritime Sciences, Chalmers University of Technology},
addressline={Chalmersgatan 4}, 
city={Gothenburg},
postcode={412 96}, 
country={Sweden}}

\affiliation[inst3]{organization={Department of Computer Science and Engineering, Chalmers University of Technology},
            addressline={Chalmersgatan 4}, 
            city={Gothenburg},
            postcode={412 96}, 
            country={Sweden}}


\begin{abstract}
Free-text crash narratives recorded in real-world crash databases have been shown to play a significant role in improving traffic safety. However,   large-scale analyses remain difficult to implement as there are no documented  tools that can batch process the unstructured, non standardized text content written by various authors with diverse experience and attention to detail. In recent years, Transformer-based pre-trained language models (PLMs), such as Bidirectional Encoder Representations from Transformers (BERT) and large language models (LLMs), have demonstrated strong capabilities across various natural language processing tasks.
These models can extract explicit facts from crash narratives, but their performance declines on inference-heavy tasks in, for example, \textit{Crash Type} identification, which can involve nearly 100 categories. 
Moreover, relying on closed LLMs through external APIs raises privacy concerns for sensitive crash data. Additionally, these black-box tools often underperform due to limited domain knowledge.
Motivated by these challenges, we study whether compact open-source PLMs can support reasoning-intensive extraction from crash narratives. We target two challenging objectives: 1) identifying the \textit{Manner of Collision} for a crash, and 2) \textit{Crash Type} for each vehicle involved in the crash event from real-world crash narratives. To bridge domain gaps, we apply fine-tuning techniques to inject task-specific knowledge to LLMs with Low-Rank Adaption (LoRA) and BERT. Experiments on the authoritative real-world dataset Crash Investigation Sampling System (CISS) demonstrate that our fine-tuned compact models outperform strong closed LLMs, such as GPT-4o, while requiring only minimal training resources. Further analysis reveals that the fine-tuned PLMs can capture richer narrative details and even correct some mislabeled annotations in the dataset. Our code and data are publicly available\footnote{\url{https://github.com/hiXixi66/PLMs-Crash-Narrative-Analysis}}.
\end{abstract}










\end{frontmatter}



\section{Introduction}
\label{sec1}
Today, traffic crashes remain one of the leading causes of death worldwide, with an estimated 1.19 million fatalities annually~\citep{who_2023_globalstatus}. Improving road safety is therefore a critical global challenge. A key approach is through analyzing high quality real-world data recorded in crash databases~\citep{imprialou2019crash,xie2019use}. Data is typically documented by crash investigators as part of official reports and usually contains information about the crash, those involved, crash severity outcomes (i.e.,  injuries or fatalities), and crash narratives. Among these records, crash narratives are especially valuable, as they include actual descriptions of the crash that go beyond structured variables, such as vehicle travel directions, impact points, and other contextual factors~\citep{10586010}. Despite their value, crash narratives are currently limited to smaller case studies. They are typically written in unstructured text with highly diverse writing styles, inconsistent terminology, and varying levels of detail. As a result, these narratives cannot be readily batch-processed using conventional statistical techniques. Investigators have traditionally addressed this by manually preprocessing the text into structured formats such as tables for statistical analysis. This process is not only labor-intensive but also error-prone, especially when working with large datasets.

To reduce this burden, researchers have increasingly explored natural language processing (NLP) to automate parts of the pipeline. Early approaches relied on topic modeling (e.g., Latent Dirichlet Allocation, LDA~\citep{blei2003latent}), Term Frequency-Inverse Document Frequency (TF-IDF)~\citep{christian2016single} keyword weighting, or bag-of-words (BoW)~\citep{zhang2010understanding} classifiers. While useful for coarse crash classification, these methods process each word separately and cannot capture sentence structure or nuanced semantics in a crash narrative. Word embedding methods improved semantic representation by mapping words to vectors, and recurrent neural networks (RNNs) further advanced sequence modeling by learning from data. In practice, these techniques provide only shallow semantic cues, which can extract explicit information from narratives but remain insufficient for complex crash narratives where multiple vehicles are referenced in a single text description which might include crash causal links. For example, collision type is an essential factor in crash analysis, but it can only be determined through in-depth interpretation of a crash narrative. Extracting the \textit{Crash Type} for each vehicle is even more challenging. In multi-vehicle crashes, the crash report is often written as a single narrative where descriptions of different vehicles are intertwined and sometimes causally related. In such cases, models must accurately identify the target vehicle and make decisions based on its context without being distracted by irrelevant textual information, a capability that the  previously mentioned models struggle to provide.

Transformer-based pre-trained language models (PLMs), like Bidirectional Encoder Representations from Transformers (BERT)~\citep{devlin2019bert} and Large Language Models (LLMs), provide better solutions \citep{vaswani2017attention}. By pre-training on large-scale corpora, models learn to capture rich semantic representations, thereby gaining powerful sentence understanding capabilities. BERT is suitable for classification tasks because its pre-training on masked language modeling, but it relies on patterns learned from training data, without considering the semantic meaning of categories or incorporating any oracle knowledge. In contrast, LLMs can exploit such knowledge, and techniques like prompt engineering, chain-of-thought (CoT)~\citep{wei2022chain} prompting, and few-shot learning are widely used to further enhance their performance. 

Yet, despite their strong capabilities, applying PLMs to crash narrative analysis faces several challenges. First, the use of external APIs may introduce data security and privacy risks, as sensitive crash reports cannot always be shared with third-party services. Second, deploying and running large-scale models requires substantial computational resources, making both training and inference costly. Finally, PLMs have shown limited adaptability to traffic safety domains because their training corpora typically lack sufficient specific knowledge related to traffic.

Considering all the above, we applied fine-tuned open-source PLMs on domain-specific data to achieve inference-intensive traffic crash information extraction. Here, we reformulate the information extraction from narratives problem as a structured classification task. For straightforward classification (e.g., per-crash \textit{Manner of Collision} extraction), we fine-tune compact PLMs (e.g., BERT) on crash data to focus on safety-critical cues in narratives. For the more complex task (e.g., per-vehicle \textit{Crash Type} extraction), we use an instruction-following LLM and inject oracle knowledge (e.g., its possible category set) directly into the prompt, ensuring predictions are consistent with dataset coding rules. To mitigate data quality and quantity requirements, we apply Low-Rank Adaptation (LoRA)~\citep{hu2022lora} to parameters-efficiently adapt of the open-source LLMs.
We evaluate our approach on the Crash Investigation Sampling System (CISS) dataset released by the U.S. National Highway Traffic Safety Administration (NHTSA)~\citep{nhtsa2023ciss}, comparing the results against GPT models, BERT, and seven LLM backbones as baselines.

Our contributions are threefold:
\begin{itemize}
    \item First, we formulate crash-narrative understanding as a structured classification problem and demonstrate that domain-adapted PLMs can effectively solve it. 
    \item Second, we enable constraint-aware, fine-grained \textit{Crash Type} classification by instruction-tuning LLMs to explicitly respect label-space restrictions imposed by upstream category and configuration choices.
    \item Finally, we provide a comprehensive empirical study on CISS, showing state-of-the-art performance, data efficiency, and robustness.
\end{itemize}

\section{Related work}
\subsection{Traditional text-analysis methods}

Research on crash narrative analysis typically follows a two-stage pipeline. First extracting text features (e.g., frequencies, topics, embeddings) and then applying classification models. Early work relied on frequency-based representations. BoW~\citep{zhang2010understanding} treats a text as an unordered collection of words and encodes text as simple word counts or TF-IDF values. These representations are typically paired with traditional classifiers such as Naive Bayes~\citep{aborisade2018classification}, Logistic Regression~\citep{aborisade2018classification,akuma2022comparing}, and Support Vector Machines (SVMs)~\citep{cichosz2018case}. Such models have been used, for example, to identify agricultural crashes~\citep{kim2021crash} and secondary crashes~\citep{zhang2020identifying}.
However, BoW and TF-IDF only reflect surface-level word frequency statistics and fail to capture the latent semantic structure of documents. To incorporate semantic themes, researchers adopted topic modeling.
Methods such as LDA learn latent topics from text and represent each narrative as a probability distribution over these topics. Some studies have applied topic modeling to summarize themes in travel surveys~\citep{baburajan2018opening, baburajan2020open}, to study motorcycle crash causation~\citep{das2021topic}, to analyze the relationships between latent topics and crash severity~\citep{li2024analyzing}, and to examine safety concerns associated with transitions of control in autonomous vehicle crash narratives~\citep{alambeigi2020crash}. crashes~\citep{alambeigi2020crash}.
Yet, topic models remain coarse-grained, providing only distributions over topic words and lacking detailed semantic understanding. With the rise of word embeddings, text representation have advanced beyond sparse counts.
Techniques such as Word2Vec~\citep{mikolov2013efficient} and GloVe~\citep{pennington2014glove} project words into continuous low-dimensional spaces, capturing semantic and syntactic similarities. When integrated with downstream classifiers, these embeddings have been used for traffic text classification to detect emerging risks~\citep{kim2020word} or for transportation sentiment analysis~\citep{ali2019transportation}. To further improve performance, neural networks such as CNNs~\citep{qiao2022construction} and RNNs~\citep{heidarysafa2018analysis} have been employed as advanced downstream classifiers following word embeddings. For example, some researchers used them in railway accident cause analysis~\citep{heidarysafa2018analysis}, road crash cause analysis~\citep{xiong2024crash}, and injury outcomes investigation of Horse-and-Buggy crashes in rural areas~\citep{qawasmeh2024investigating}.

The above methods decouple text representation from classification, performing feature extraction and task learning independently, preventing gradients from back-propagating through the entire model during training. As a result, the text representations cannot be optimized for the specific task, which limits the model's overall performance. Recent research has shifted toward unified models that integrate text representation and downstream task learning in an end-to-end manner. For instance, some studies learn richer semantic features for identifying accident causes~\citep{zhang2022hybrid} or performing traffic classification~\citep{ren2021tree}. However, RNNs still suffer from gradient vanishing and gradient exploding problems, which restrict layer stacking beyond a few layers and limit their ability to capture complex semantics.

\subsection{PLM-based text-analysis methods}
With the rise of Transformer architectures, model stacking is no longer a limitation, and model parameters can and have scaled to billions. This scaling has significantly enhanced model capacity, enabling the use of large-scale pretraining on massive corpora to capture contextual semantic representations. For example, BERT has been applied to identify actual wrong-way driving (WWD) crashes~\citep{hosseini2023application}, to classify traffic injury types into five categories~\citep{oliaee2023using}, and to extract impact points and pre-collision vehicle maneuvers~\citep{seo2023text}. However, BERT is pre-trained with masked language modeling and designed for classification instead of prompt-driven generation, so it cannot directly interpret and respond to prompts, limiting its applicability in crash narrative analysis. 

In contrast, prompt-based LLMs can naturally incorporate task instructions into the input, making them more flexible for classification and reasoning over unstructured crash narratives. Mumtarin et al.~\citep{mumtarin2023large} evaluated LLMs in answering safety question tasks from accident narratives by embedding questions directly into prompts. Arteaga and Park~\citep{arteaga2025large} employed prompt engineering to identify unreported alcohol involvement in crash reports. Zhen et al.~\citep{zhen2024leveraging} further leveraged CoT~\citep{wei2022chain} reasoning and prompt engineering with LLMs to enhance traffic crash severity analysis and inference. To address the limited adaptability of PLMs caused by insufficient domain-specific knowledge in their training corpora, fine-tuning Transformer-based models provides an effective solution. For example, Jaradat et al.~\citep{jaradat2024multitask} proposed
a multi-task learning (MTL) LLM framework, and Golshan et al.~\citep{golshan5379743extracting} fine-tuned the LLaMA 3.1 model to extract crash locations and casualty counts, with both approaches achieving high accuracy.

Overall, these studies focus on simple tasks with few categories or information explicitly stated in the narratives, leaving their effectiveness on more complex problems unknown.

\section{Methodology}

PLMs are effective for text classification, converting unstructured crash narratives into structured fields~\citep{harne2025llm}. While recent traffic-safety studies can extract explicitly stated details~\citep{du2023safety}, reasoning-intensive variables remain challenging due to limited crash-specific knowledge and the prohibitive cost of full-model training. We address this by adopting parameter-efficient fine-tuning on BERT and with LoRA~\citep{hu2022lora} on LLMs using an annotated crash dataset, injecting domain knowledge at low compute and memory cost and enabling safer, more controllable deployment than closed LLMs via public APIs. Our work targets two high-value variables that support crash risk assessment, and countermeasure design: (i) \emph{\textit{Manner of Collision}}, a reasoning task over event sequences and interacting agents; and (ii) \emph{Crash Type}, a fine-grained classification with 98 possible classes. We now begin by describing the data used in our approach, defining our problem more clearly for both tasks, and presenting our fine-tuning to extract information from crash narratives.

\subsection{Dataset: CISS}
\label{sec:dataset}
This paper uses the Crash Investigation Sampling System (CISS) dataset released by the U.S. National Highway Traffic Safety Administration (NHTSA)~\citep{nhtsa2023ciss} as a case study. CISS is a national traffic crash database which provides comprehensive traffic crash data on police-reported motor vehicle crashes occurring in the United States involving passenger cars, light trucks, and light vans that were towed~\citep{radja2022crash}. It retrospectively investigated traffic crash records collected from 2017 to 2023, of approximately 3,700 cases per year. The database contains about 39 relational tables\footnote{The number and columns of tables in CISS continuously changes as the data-collection process is  updated. For instance, the dataset of year 2017 did not include the \texttt{VPICDECODE} table.}, each describing a specific aspect of the crash, such as crash level data (table \texttt{CRASH}), events level record (table \texttt{EVENT}), general vehicle record (table \texttt{GV}). CISS includes not only \texttt{SUMMARY} (a basic description of the crash scenario as documented by the crash investigator), but also structured case coding such as \texttt{MANCOLL} (manner of collision) and \texttt{CRASHTYPE} (crash type), which are basically manually determined~\citep{national2023nhtsa}. 

In the CISS dataset, \texttt{MANCOLL} is not directly annotated but rather derived through rule-based mapping from other manually labeled variables, including \texttt{OBJECT CONTACTED} in table \texttt{EVENT} as well as \texttt{CRASHTYPE} and \texttt{TRANSPORT} in table \texttt{GV}. The resulting classification consists of seven categories as shown in Table~\ref{tab:mancoll-classes}, one of which corresponds to unknown.

\begin{figure}
    \centering
    \includegraphics[width=0.95\linewidth]{ 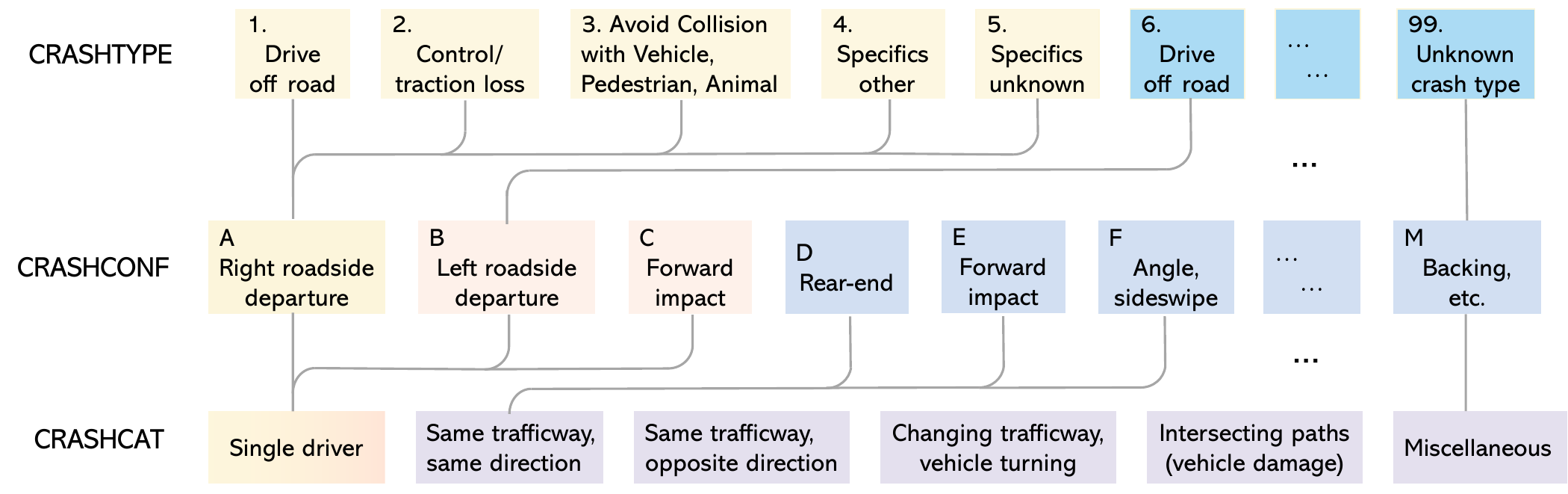}
    \caption{Hierarchical mapping of \texttt{CRASHTYPE} in a two-step classification procedure}
    \label{fig:crash-type-conf-cat}
\end{figure}

\begin{table}[ht]
\footnotesize
\centering
\caption{Class definitions for the \texttt{MANCOLL} task.}
\begin{tabular}{cl}
\toprule
\textbf{Label ID} & \textbf{MANCOLL} \\
\midrule
0 & Not Collision with Vehicle in Transport \\
1 & Rear-End \\
2 & Head-On \\
4 & Angle \\
5 & Sideswipe, Same Direction \\
6 & Sideswipe, Opposite Direction \\
9 & Unknown \\
\bottomrule
\end{tabular}
\label{tab:mancoll-classes}
\end{table}
 The \texttt{CRASHTYPE} is a numeric value derived through a two-step process: first by selecting the crash category (\texttt{CRASHCAT}) and crash configuration (\texttt{CRASHCONF}) in the \texttt{GV} table (as illustrated in Figure~\ref{fig:crash-type-conf-cat}), and then by the crash investigator’s assessment based on police reports, scene inspections, vehicle inspections, and interviews.  This two-step procedure is preferred because it provides a more structured and interpretable way to visualize crash scenarios.

\subsection{Problem definition}

This paper uses the CISS dataset as a case study to evaluate BERT and LLMs on extracting information that requires deep inference from unstructured text. We focus on analyzing the crash narratives (the \texttt{SUMMARY} column in the \texttt{CRASH} table) to infer structured collision-type information. Specifically, our goal is (1) to identify the \textit{Manner of Collision} for each crash and (2) the \textit{Crash Type} for all vehicles involved in a crash. 

For the first task, we are interested in assigning the \textit{Manner of Collision} based solely on the textual description of the crash (i.e., \texttt{SUMMARY} column in the \texttt{CRASH} table), without using any structured metadata. It requires deep-reasoning beyond explicitly stated facts.
We use the \texttt{MANCOLL} column in \texttt{CRASH} as ground truth labels for accuracy evaluation.

For the second task, we want to extract the \textit{Crash Type} for each vehicle in a crash. Unlike \texttt{MANCOLL}, which involves only seven classes, \texttt{CRASHTYPE} is considerably more challenging, encompassing 97 fine-grained categories. To make this problem more tractable, we exploit the hierarchical mapping illustrated in Figure~\ref{fig:crash-type-conf-cat} and decompose the task into 13 smaller classification subtasks based on \texttt{CRASHCONF}. All subtasks are addressed within a single model. Since \texttt{CRASHCAT} and \texttt{CRASHCONF} classifications are relatively straightforward\footnote{The \texttt{CRASHCAT} classification is very similar to \texttt{MANCOLL}, consisting of only six categories. After completing the first task, we found the two tasks highly overlapping and therefore did not pursue \texttt{CRASHCAT} further. The \texttt{CRASHCONF} categories are derived from \texttt{CRASHCAT}, with each \texttt{CRASHCAT} corresponding to only 1–3 \texttt{CRASHCONF} classes, making it a relatively simple classification task as well.}, we treat \texttt{CRASHCONF} as oracle knowledge and concentrate our analysis on the more difficult \texttt{CRASHTYPE} classification.

\subsection{LLM Fine-tuning Workflow}
\label{sec:llm-fine-tuning}
As mentioned before, we formulate the information extraction task as a classification problem by prompting the LLM to output a predefined label rather than free-form text and focus on per-crash \textit{Manner of Collision} and per-vehicle \textit{Crash Type} extraction tasks. We now outline the detailed processes of the two tasks, followed by the supervised fine-tuning process. 

\begin{figure}[!t]
\centering
\subfloat[Per-crash \textit{Manner of Collision} extraction process]{
\includegraphics[width=0.99\linewidth]{ 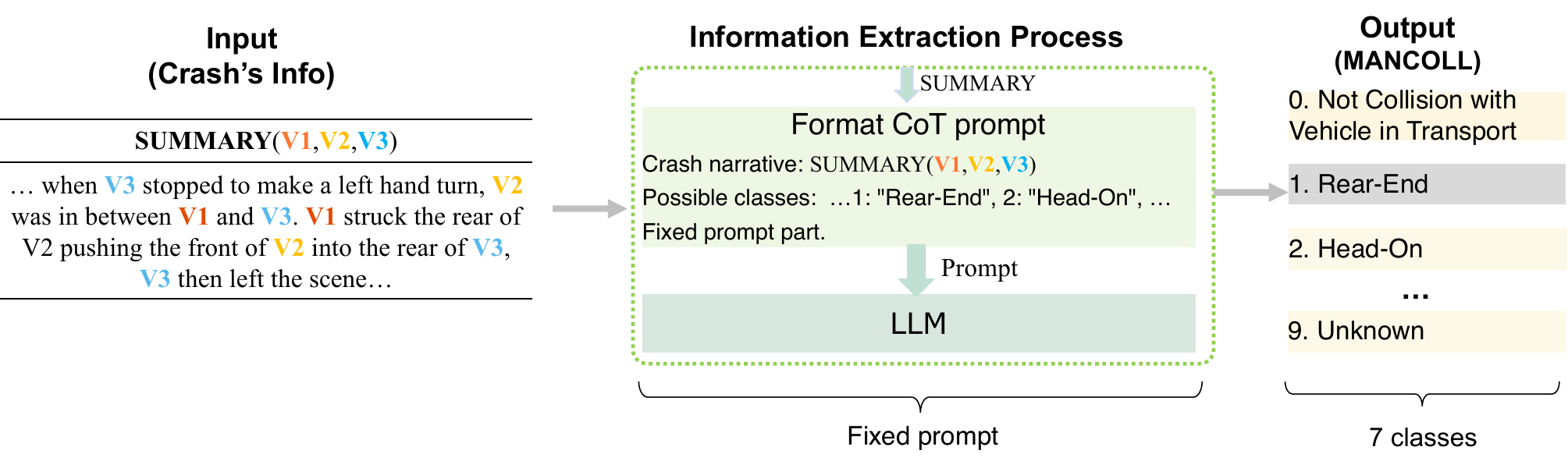}
\label{fig:mancoll-framework}}
\hfil
\subfloat[Per-vehicle \textit{Crash Type} extraction process]{
\includegraphics[width=0.99\linewidth]{ 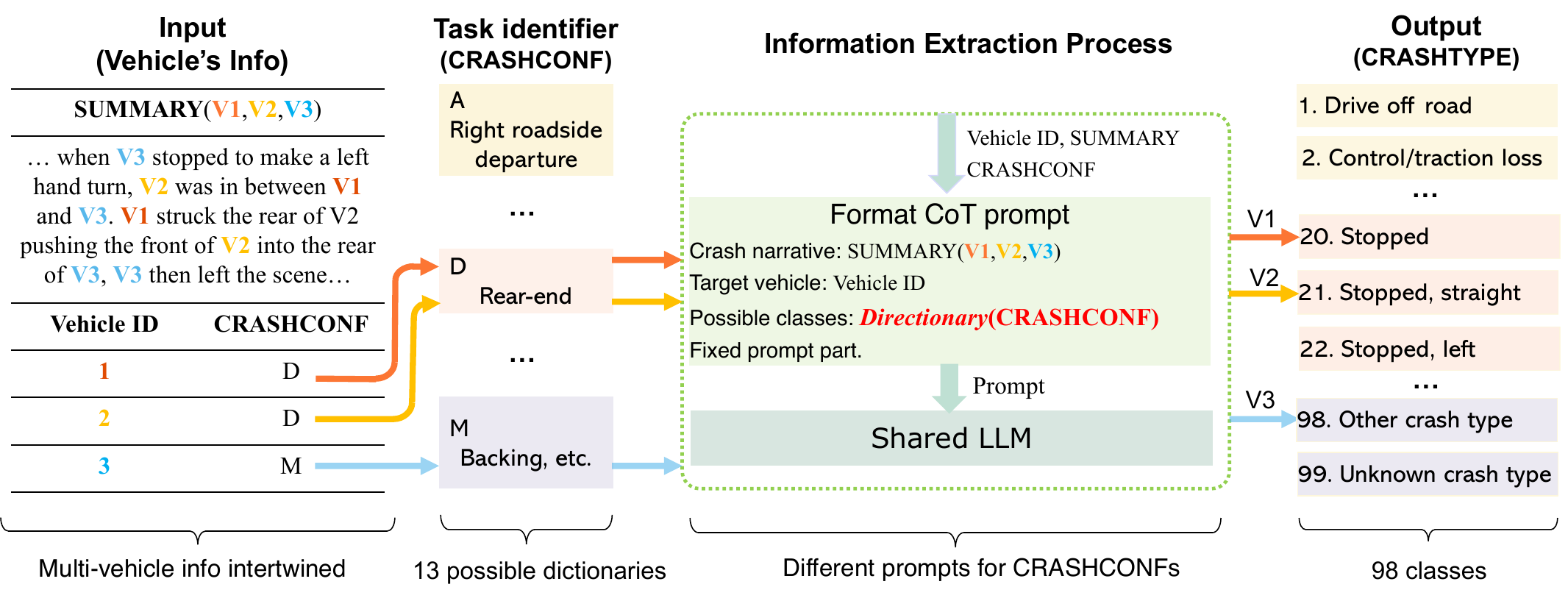}
\label{fig:crashtype-framework}}
\caption{Overview of the two information extraction tasks: (a) Manner of collision extracted from the crash narrative with a fixed CoT prompt and a small candidate set (7 classes). (b) Per-vehicle \textit{Crash Type} extracted from an intertwined multi-vehicle narrative, using 13 task-specific prompts to predict among 98 fine-grained classes.}
\label{fig:tasks-all}
\end{figure}

The information extraction and reasoning tasks involved in this paper are shown in Figure~\ref{fig:tasks-all}. In Figure~\ref{fig:mancoll-framework}, \texttt{MANCOLL} directly determines the overall collision mode from the entire crash narrative. 
The output space is relatively small (7 classes) and can be completed using a fixed CoT prompt. \texttt{CRASHTYPE} extraction for each vehicle, in Figure~\ref{fig:crashtype-framework}, requires first identifying the target vehicle \texttt{(V1/V2/V3)} within a text containing multiple vehicle information intertwined. Based on its crash configuration identifier (\texttt{CRASHCONF}), the corresponding subtask is selected from 13 different dictionaries, and the corresponding 98 \textit{Crash Types} are output. In contrast, \texttt{CRASHTYPE} classification not only involves entity and reference resolution, but also requires discrimination within a heterogeneous and larger label space. The prompts must also be dynamically constructed rather than fixed. Therefore, it is significantly more difficult to understand than Task 1.

\begin{figure}
    \centering
    \includegraphics[width=0.95\linewidth]{ 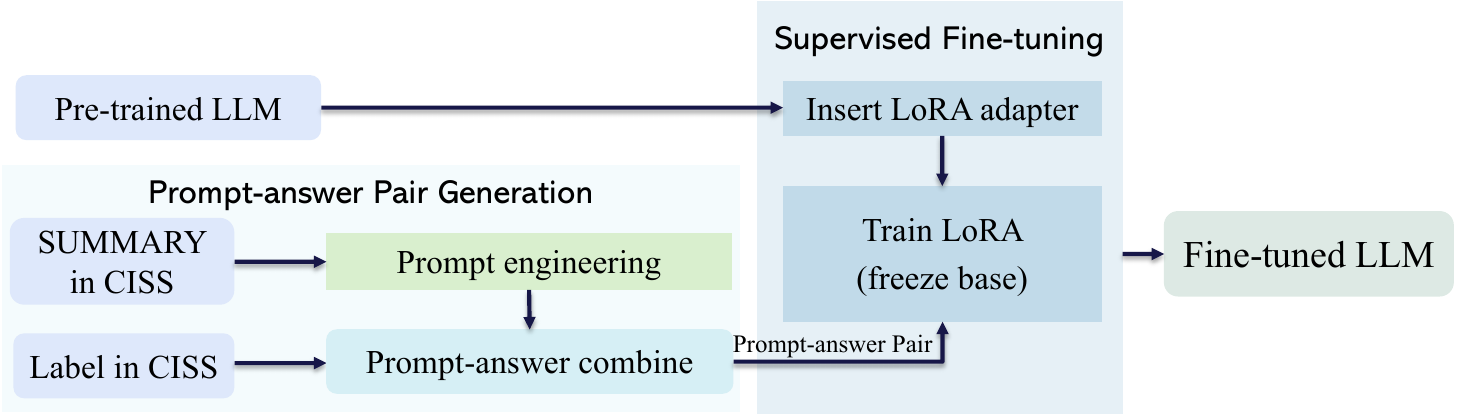}
    \caption{Overview of the proposed LoRA-based fine-tuning framework.}
    \label{fig:framework}
\end{figure}

To adapt LLMs to traffic-safety tasks, we fine-tune open-source models following the pipeline in Figure~\ref{fig:framework}. 
During fine-tuning, we first construct prompt–answer training pairs using the prompts generated in Figure~\ref{fig:tasks-all} and the labels from CISS. We then insert LoRA adapters and apply parameter-efficient fine-tuning (PEFT) to train only the adapters, reducing computational overhead and speeding up training. Finally, we obtain the fine-tuned model for inference by merging the trained adapter weights with the base model’s pre-trained weights.

\subsubsection*{Prompt engineering}
The prompt is crucial for LLM performance, as small changes in wording can greatly affect accuracy. For crash-narrative analysis, prompts must embed traffic safety knowledge to ensure reliable predictions. Accordingly, we design a structured CoT prompt with clearly separated variable slots and a fixed instruction block (Figure~\ref{fig:tasks-all}): (i) Crash narrative: \texttt{SUMMARY(V1,V2,V3)}; (ii) Target vehicle: Vehicle ID (used for \texttt{CRASHTYPE}); (iii) Possible classes: fixed small set for \texttt{MANCOLL}, but dictionary-specific for \texttt{CRASHTYPE}; and (iv) a fixed prompt part, which steers the model to understand the task, reason step-by-step, and produce constrained outputs. The prompt contains mainly the following components:

(1) Task Introduction. This part contains introduction of task (fixed prompt part) and expert-verified definitions for each category.
\begin{verbatim}
You are a helpful assistant that classifies vehicle collisions
into one of the following categories based on...
< Possible classes and their definitions >
\end{verbatim}

(2) Clarification Rules. These clarification rules are derived from domain-specific heuristics, codified through expert consultation and data observation. We first enhance model performance by applying a CoT strategy. Specifically, the prompts are designed to decompose complex reasoning tasks into step-by-step subproblems, thereby guiding the LLM to follow an explicit reasoning trajectory before arriving at the final categorical prediction. Other rules were distilled from empirical insights obtained via manual inspection and a pilot study over some training examples. For the \texttt{MANCOLL} task, this portion of the prompt is fixed, while for the \texttt{CRASHTYPE} task, it is dictionary-specific and thus varies across subtasks.

(3) Output Instruction. LLMs may produce varied outputs even when the classification result is correct, e.g., returning "4", "angle", or "angled side impact" for the same class. Such variations hinder batch processing. To avoid this, we use fixed prompt and
explicitly instruct the model to output only a valid class index from the predefined label space.
\begin{verbatim}
Only respond with a single number from the list above.
Do not add any explanation.
\end{verbatim}
All prompts used in this work are included in the~\ref{prompt-all}.

\subsubsection*{LoRA Adapter}
\begin{figure}
    \centering

    \includegraphics[width=1\linewidth]{ 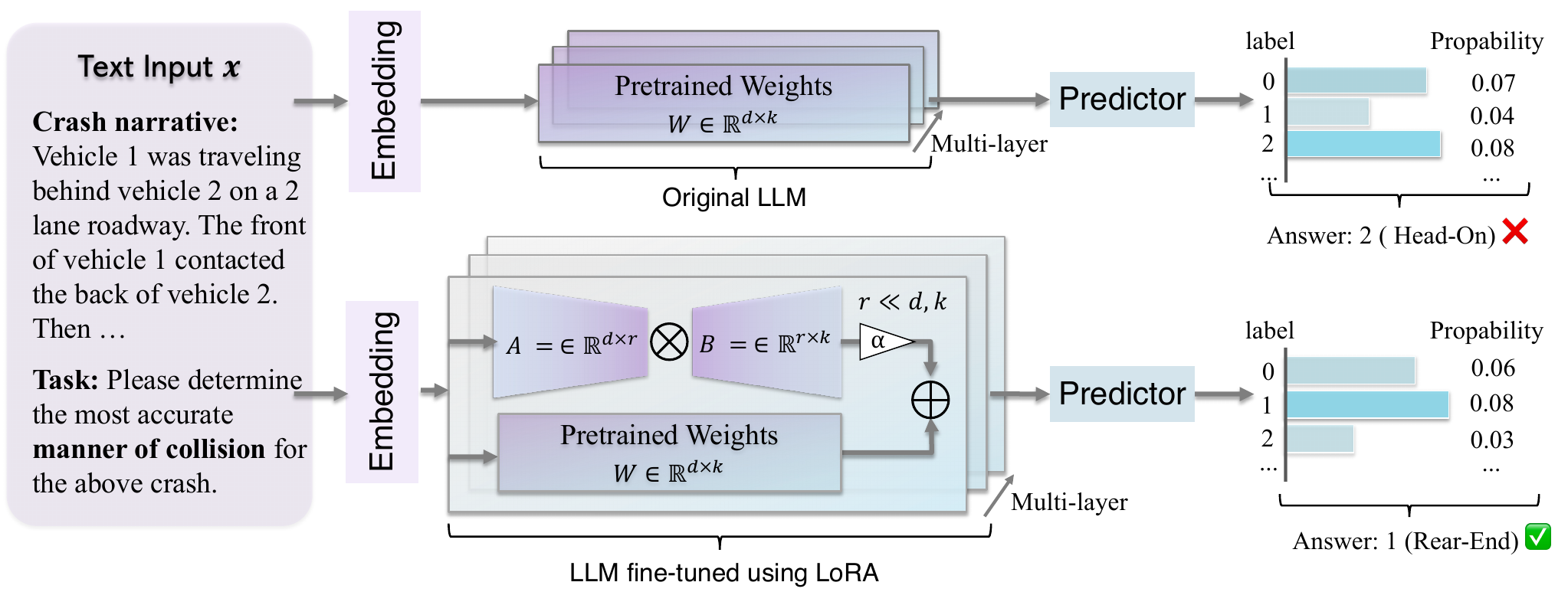}
    \caption{Prediction processes of the original LLM (top) and the LoRA fine-tuned LLM (bottom).}
    \label{fig:lora}
\end{figure}

The foundation of modern LLMs is the Transformer architecture, whose core component is the self-attention mechanism~\citep{vaswani2017attention}. By scaling up this architecture to billions of parameters and training on massive text corpora, LLMs acquire broad linguistic and world knowledge. However, traffic-domain narratives (e.g., crash reports, incident logs, work-zone descriptions) are underrepresented in such corpora, so pre-trained LLMs struggle to capture domain-specific entities, relations, and causal patterns critical for crash narrative analysis. To bridge this gap efficiently, we adopt parameter-efficient fine-tuning by keeping the pretrained backbone frozen and learning lightweight adapters specialized for crash narratives analysis.

\paragraph{Self-Attention Primer (Where Adaptation Can Act)}
Given a token sequence represented by hidden states $X \in \mathbb{R}^{T \times d}$, a single attention head computes
\[
Q = X W_Q,\quad K = X W_K,\quad V = X W_V,
\]
where $Q$, $K$, $V$ denote the query, key, and value representations obtained through the projection matrices $W_Q$, $W_K$, and $W_V$, respectively. The attention output is
\begin{equation}
\text{Attention}(Q, K, V) = \mathrm{softmax}\!\left(\frac{QK^{\top}}{\sqrt{k}}\right)V,
\label{equ:attention}
\end{equation}
This decomposition exposes three linear projections, $W_Q, W_K, W_V$, as natural adaptation points.
For traffic narratives, adjusting these projections helps the model aligns with crash-specific cues, and emphasize values relevant to causality.

\paragraph{LoRA for Parameter-Efficient Domain Adaptation}
Low-Rank Adaptation (LoRA)~\citep{hu2022lora} inserts trainable low-rank matrices into selected frozen projections so that only a small number of parameters are updated. A comparison between the original LLM and the LoRA fine-tuned LLM is illustrated in Figure \ref{fig:lora}.  For the same input, the original model fails to produce the correct answer, whereas the fine-tuned model adapts the LLM's original weights $\textbf{W}$ by adding the product of low-rank matrices $\textbf{A}$ and $\textbf{B}$, which alters the label prediction probabilities and enables the correct output. 

Specifically, for a target weight $W \in \mathbb{R}^{d \times k}$ (e.g., one of $W_Q, W_K$ or $W_V$), LoRA learns a low-rank update
\[
\Delta W = \frac{\alpha}{r}\, A B,\qquad A \in \mathbb{R}^{d \times r},\; B \in \mathbb{R}^{r \times k},\; r \ll \min(d,k),
\]
and uses the adapted weight
\begin{equation}
W' = W + \Delta W
\end{equation}
During fine-tuning, the large pretrained $W$ is \emph{frozen} and only $A,B$ are trained. 
This reduces trainable parameters from $d\times k$ to $(d+k)r$ and lowers memory cost, while preserving general linguistic competence. 
Crucially, because attention is factorized into $Q,K,V$, placing LoRA on $W_Q$/$W_K$ directly modulates the \emph{attention weights} in Equation~\eqref{equ:attention}, enabling the model to learn traffic-specific matching patterns; placing LoRA on $W_V$ refines \emph{content aggregation and readout}, improving how incident attributes and causal chains are represented downstream.
In our implementation, we apply LoRA to all the query, value projection layers, and key projection layers ($W_Q, W_K$ and $W_V$) of the transformer blocks, which are empirically found to be effective for language modeling tasks \citep{mao2025survey}.

\paragraph{Loss function} In our tasks, each crash or vehicle is assigned a label corresponding to the predefined classes. To simplify the output, we encode all class labels as single-token numeric identifiers using a dictionary mapping, such as in Table~\ref{tab:mancoll-classes}. Consequently, the model is trained to generate a single token representing the class ID for each input.

We adopt the standard cross-entropy loss for training because it directly measures the divergence between the predicted probability distribution over the vocabulary and the one-hot target distribution~\citep{zhang2018generalized}, thereby encouraging the model to assign maximal probability to the correct class token. Formally, let \textit{x} denote the input sequence (i.e., the crash summary), and let $y \in \{0, 1, ..., k\}$ be the target class index, where \textit{k} is the total number of classes. At the final decoding step, the LLM outputs a vocabulary-sized logit vector $\mathbf{z} \in \mathbb{R}^{|V|}$. The probability of generating the target class token is computed via the softmax function:

\begin{equation}
P(y \mid x) = \frac{\exp(z_y)}{\sum_{i=1}^{|V|} \exp(z_i)}
\end{equation}

The training objective is to minimize the negative log-likelihood of the correct class token, which corresponds to the standard cross-entropy loss:

\begin{equation}
\mathcal{L}_{\text{CE}} = -\log P(y \mid x) = -\log \left( \frac{\exp(z_y)}{\sum_{i=1}^{|V|} \exp(z_i)} \right)
\end{equation}

\subsection{BERT Fine-tuning}
To adapt BERT for classification tasks, a linear layer is added on top of the BERT encoder. Specifically, the hidden representation of the special \texttt{[CLS]} token is used as the sentence embedding $h_{[CLS]}$. The prediction is obtained as
\begin{equation}
\hat{y} = \text{softmax}(W \cdot h_{[CLS]} + b),
\end{equation}
where $W$ and $b$ are trainable parameters of the linear classifier. During fine-tuning, the objective is to minimize the cross-entropy loss between $\hat{y}$ and the ground-truth label, identical to the loss function used as LLM fine-tuning in Section~\ref{sec:llm-fine-tuning}. Importantly, all parameters of BERT together with the classification layer are updated end-to-end, enabling the model to learn task-specific knowledge while retaining its pretrained linguistic representations.

\section{Experiments}
\subsection{Experimental Settings}

\textbf{Backbones and baselines} We evaluated BERT and multiple LLMs on the classification task under different fine-tuning configurations. The backbone PLMs include several frontier open-source families: BERT~\citep{devlin2019bert}, LLaMA3 series~\citep{meta2024llama3} (LLaMA3.2-1B, LLaMA3.2-3B, LLaMA3.1-8B, LLaMA3.3-70B), Qwen series~\citep{qwen2.5} (Qwen2.5-7B-Instruct), and Mistral series~\citep{jiang2024mixtral} (Mistral-7B-Instruct-v0.3). We also evaluated the closed model GPT-4o~\citep{openai2024gpt4o},  recurrent neural network baseline like TextRNN~\citep{liu2016recurrent} and FastText~\citep{joulin2016bag}. Parameter size and deployment resources of PLMs are shown in Table~\ref{tab:model-comparison}\footnote{Model sizes and deployment requirements are referenced from the official NVIDIA NIM documentation: \url{https://docs.nvidia.com/nim/large-language-models/latest/introduction.html}.}. To ensure a fair comparison, all experiments were conducted on a single NVIDIA A100 GPU, except for LLaMA3-70B, which required 4 A100 GPUs due to its larger size.

\begin{table}[ht]
\footnotesize
\centering
\caption{Overview of backbone models: parameter size and deployment resources.}
\setlength{\tabcolsep}{10pt} 

\begin{tabular}{lcccc}
\toprule
\textbf{Model} & \textbf{Parameters} & \textbf{GPU Memory Required}  \\
\midrule
BERT     & $\approx$ 0.11 B           & $\approx$  0.23GB\\
\hline
\multicolumn{3}{l}{\textit{Open-source LLMs}}\\
\hline
LLaMA3.2-1B      & $\approx$ 1.23 B           & $\approx$ 3 GB \\
LLaMA3.2-3B      & $\approx$ 3.21 B         & $\approx$ 6 GB \\

Qwen2.5-7B-Instruct      & $\approx$ 7 B          & $\approx$ 14 GB  \\
Mistral-7B-Instruct-v0.3     & $\approx$ 7.30 B   & $\approx$ 14 GB \\
LLaMA3.1-8B      & $\approx$ 8 B          & $\approx$ 15 GB  \\
LLaMA3.3-70B-Instruct     & $\approx$ 70 B          &  $\approx$ 131 GB \\
\hline
\multicolumn{3}{l}{\textit{Closed LLMs}}\\
\hline
GPT-4o  & –-      & –-             \\
\bottomrule
\end{tabular}
\label{tab:model-comparison}
\end{table}

\textbf{Dataset} To assess the generalizability of our approach to future police reports, model training was performed on crash data of year 2020, consisting of 300 annotated examples for prompt design and adjustment and about 3,000 examples for fine-tuning. We evaluated on 2,000 test cases from 2021 to measure inference performance.

\textbf{Hyper-Parameters} Inference was conducted with a fixed temperature of 0.2, controlling the randomness of output generation. Through multiple pilot experiments, we found that this value offered the best trade-off between accuracy and robustness. LLMs are fine-tuned with LoRA, whereas BERT is fully fine-tuned. Following \citep{raschka2024practical}, we set $r=8$ and $\text{LoRA}_\alpha=16$, which balance training efficiency and performance for the classification task, providing sufficient task-specific adaptation without excessive computational cost.  Our fine-tuning and test scripts are publicly available\footnote{\url{https://github.com/hiXixi66/PLMs-Crash-Narrative-Analysis}}.


\subsection{Manner of collision}
To evaluate both the overall correctness of predictions and the balance of performance across different classes, we assessed model performance using accuracy and Macro F1-score~\citep{sokolova2009systematic}. In addition, deploying, training, and running inference with LLMs and BERT require substantial computational resources. To better understand the trade-offs, we compare models of different sizes and from different providers, reporting both training and inference times. These results are then analyzed together with accuracy and Macro F1-score to provide a comprehensive evaluation of the whole framework.

During manual inspection, we found that \textit{Unknown} (label ID: 9) in the original dataset could reasonably be reassigned to other valid categories. Therefore, in addition to reporting results on the original dataset, we also present evaluations with this label \textit{Unknown} excluded to provide a more precise assessment of model performance.

\subsubsection*{Results of different baselines and LLMs on \textit{Manner of Collision}}

The overall results of \texttt{MANCOLL} classification are summarized in Table~\ref{Main-table-of-MANCOLL}. We reported results both including and excluding \textit{Unknown} in the column \textit{ALL} and \textit{–Unknown} respectively.
\begin{table}[ht!]
\scriptsize
\centering
\caption{Results of various LLMs on Manner of Collision (\texttt{MANCOLL}) classification task.}
\setlength{\tabcolsep}{3pt} 
\renewcommand{\arraystretch}{0.8} 
\begin{tabular}{lccccccc}
\toprule
\multirow{2}{*}{\textbf{Backbones}} 
& \multirow{2}{*}{\textbf{Training step}} 
& \multirow{2}{*}{\makecell{\textbf{Training time}\\\textbf{(s)}}} 
& \multirow{2}{*}{\makecell{\textbf{Inference time}\\\textbf{(ms)}}} 
& \multicolumn{2}{c}{\textbf{Accuracy($\%$)}} & \multicolumn{2}{c}{\textbf{Macro F1}} \\
\cmidrule(l){5-6}\cmidrule(l){7-8}
& & & &  {All} & {-Unknown} & {All} & {-Unknown}\\
\midrule
\multicolumn{8}{l}{\textbf{\textit{Baselines}}} \\
\midrule
\multicolumn{1}{l}{TextRNN} &525& 0.9 &$ < $ 1.0 & \textbf{41.9} & \textbf{42.6}  & \textbf{0.093} & \textbf{0.110} \\
\midrule
\multicolumn{1}{l}{FastText}  &525& 1.1 & $ < $ 1.0& \textbf{78.8} & \textbf{80.4}  & \textbf{0.342} & \textbf{0.407} \\
\midrule
\multicolumn{8}{l}{\textbf{\textit{Open source LLMs}}} \\
\midrule
\multirow{3}{*}{BERT} 
& 169&29.8&3.9&85.3&86.7&0.429&0.507 \\
& 507&89.1&4.0&91.2&92.7&0.551&0.650 \\
&845&148.6&4.0&\textbf{92.9}&\textbf{94.3} &\textbf{0.637}&\textbf{0.620} \\

\midrule

\multirow{5}{*}{LLaMA3-1B} 
& \multicolumn{2}{c}{Original} & 25.6& 1.6 & 0.0 & 0.005 & 0.000\\
\cmidrule(l){2-8}
& 417  & 112.9 & 20.1 & 45.4 & 46.2 & 0.151 & 0.178\\
& 834 & 226.7 &21.1 & 81.9 & 83.3 & 0.359 & 0.423 \\
& 1251 & 339.8 & 20.4 & 85.8 & 87.2 & 0.407 & 0.479 \\
& 1668 & 452.4 & 20.0 & \textbf{87.5} & \textbf{88.9} & \textbf{0.491}& \textbf{0.496} \\
\midrule
\multirow{5}{*}{LLaMA3-3B} 
& \multicolumn{2}{c}{Original} & 33.8 & 33.7 & 34.9 & 0.184 & 0.143 \\
\cmidrule(l){2-8}
& 417  & 222.1 & 33.5 & 92.8 & 94.4 & 0.690 & 0.815\\
& 834 & 443.4 &35.3 & 94.0 & 95.4 & 0.754 & 0.745 \\
& 1251 & 668.9 & 33.7 & 95.0 & 96.4 & 0.762 & 0.753 \\
& 1668 & 890.5 & 33.3 & \textbf{95.1} & \textbf{96.4} & \textbf{0.779}& \textbf{0.753} \\
\midrule
\multirow{5}{*}{Qwen2.5-7B} 
& \multicolumn{2}{c}{Original} &48.2 & 76.2 & 77.0 & 0.562 & 0.559 \\
\cmidrule(l){2-8}
& 417  & 396.2 & 49.7 & 92.7 & 94.1 & 0.680 & 0.679 \\
& 834 & 795.3 & 49.9& 94.0 & 95.5 & 0.745 & 0.746 \\
& 1251 & 1192.7 & 49.5& \textbf{94.4} & \textbf{95.7} &\textbf{ 0.774} & \textbf{0.755} \\
& 1668 & 1583.9 &49.7 & 94.4& 95.7& 0.772& 0.753\\
\midrule
\multirow{5}{*}{Mistral-7B} 
& \multicolumn{2}{c}{Original} &{49.7} & 81.7 & 83.1 & 0.553 & 0.561 \\
\cmidrule(l){2-8}
& 417  & 377.63 & 48.8 & 90.1 & 91.5 & 0.680 & 0.679 \\
& 834 & 756.12 & 49.8& 92.4 & 93.3 & 0.765 & 0.729 \\
& 1251 & 1149.84 & 49.5& 91.2 & 92.2 & 0.740 & 0.701 \\
& 1668 & 1512.86 &49.7 & \textbf{94.4}& \textbf{95.7}& \textbf{0.772}& \textbf{0.753}\\
\midrule
\multirow{5}{*}{LLaMA3-8B} 
& \multicolumn{2}{c}{Original} &  54.9&34.3 &34.9 &0.214 &0.251 \\
\cmidrule(l){2-8}
& 417  & 406.6 & 56.8 & 94.4& 96.0&0.749 & 0.886\\
& 834 & 803.6 &56.8 & 95.2& 96.5&0.803 & 0.773\\
& 1251 & 1209.8 & 56.6&95.9 &97.1& 0.833&0.789 \\
& 1668 & 1615.4 & 57.1 & \textbf{96.1 }& \textbf{97.1 }& \textbf{ 0.848}& \textbf{0.788} \\
\midrule
\multirow{1}{*}{LLaMA3-70b} 
& \multicolumn{2}{c}{Original} & 508.3 & 91.4& 92.7& 0.692 & 0.704 \\
\midrule
\multicolumn{8}{l}{\textbf{\textit{Closed LLMs}}} \\
\midrule
\multicolumn{3}{l}{GPT-4o} & & 90.9 & 92.3 & 0.674 & 0.805 \\
\bottomrule

\end{tabular}
\label{Main-table-of-MANCOLL}
\end{table}
Fine-tuning brings significant improvements to open-source models and their accuracies are much higher than TextRNN and FastText. For example, the LLaMA3-3B model improves from an initial accuracy of 50.2\% to over 95.1\% after fine-tuning. At the same time, the performance of fine-tuned LLaMA3-3B and BERT models is comparable to that of 7B and 8B models and outperforms GPT-4o, despite being significantly smaller in model size and computational requirements. Training efficiency is also notable: convergence is achieved efficiently, requiring only 148.6 seconds for BERT, 890 seconds for the 3B model. Inference time also remains low, with all models below 8B requiring less than 60\,ms per instance. The fast training speeds of LLMs are largely attributed by our problem formulation, where the task is cast as a classification problem and the LLM only needs to generate a single token as output. Another important observation concerns the \textit{Unknown} class in the original dataset. As shown in Table~\ref{Main-table-of-MANCOLL}, removing this class yields a clear improvement in both accuracy and macro F1 score. Our sample analysis revealed that many instances labeled as \textit{Unknown} could in fact be reasonably mapped to known categories, and the employed LLMs were often able to classify them correctly.

Overall, these findings demonstrate that with only a small number of fine-tuning steps, even relatively small-scale PLMs can be effectively adapted for the \textit{Manner of Collision} classification task.

\subsubsection*{Effect of training data}
To better understand how data quality and quantity influence model performance, we examine the effects of label noise and varying amounts of training samples. These experiments provide insights into the robustness of high-performing models (FastText, BERT and LLMs) in the \texttt{MANCOLL} classification task and their ability to leverage limited or imperfect data.


\textbf{Effect of noisy data.} In real application scenarios, labeled data are influenced by the knowledge of human investigator, so the data cannot be guaranteed to be fully correct and may inevitably contain noise. However, the downstream application requires reliable predictions on clean inputs. Therefore, to evaluate the robustness of the models, we further examined their generalization by fine-tuning on noisy data and then testing on the clean dataset. 

\begin{figure}[!t]
\centering
\subfloat[]{
\includegraphics[width=2.55in]{ 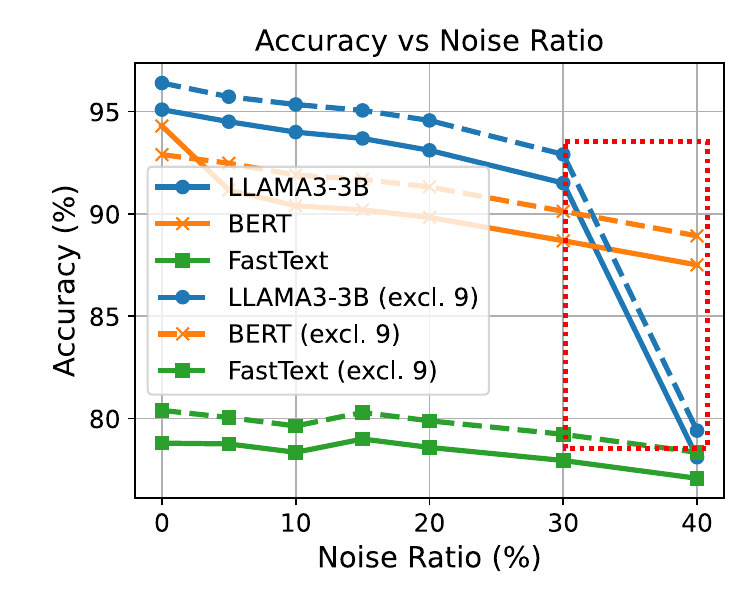}
\label{fig:image1}}
\subfloat[]{\includegraphics[width=2.55in]{ 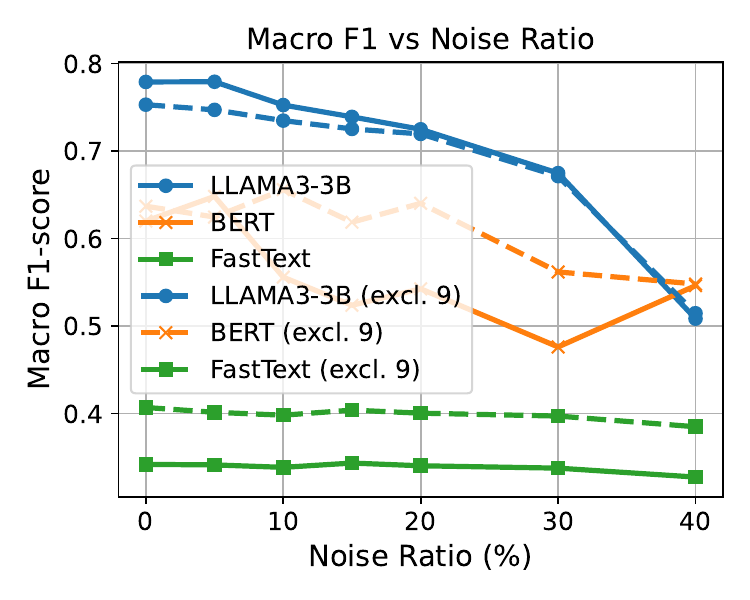}
\label{fig:image2}}
\caption{Accuracy (a) and Macro F1 (b) of LLaMA3-3B, BERT, and FastText under different noise ratios in the training data.}
\label{fig:noise}
\end{figure}

In this experiment, we introduce label noise by randomly assigning a subset of the data samples\footnote{Experiments with noise ratios above 40\% are not reported, as such levels indicate severely degraded data quality, making fine-tuning on these datasets impractical and meaningless.} to one random class, and then train models on this noisy dataset while testing on the original clean test set. As shown in Figure~\ref{fig:noise}, we observe that when the noise ratio is below 30\%, the performance of all models does not drop substantially, indicating that these methods are relatively robust to moderate levels of noise. And LLaMA3-3B consistently outperforms BERT, benefiting from background knowledge in its prompts, whereas BERT and FastText lack this advantage. Nevertheless, when the noise ratio increases further (the red box in Figure~\ref{fig:noise}a), the performance of the LLM drops sharply. This is likely due to knowledge conflicts between the prior knowledge encoded in the prompts and the noisy training signals.

\textbf{Effect of training samples amount.} To enhance the performance of pre-trained language models on traffic data analysis through fine-tuning, a sufficient amount of annotated data is required. However, producing large annotation quantities is labor-intensive and costly. Therefore, we conduct experiments to analyze how the size of the training dataset affects model performance. Out of this concern, we investigate the impact of the number of training samples on model performance.

\begin{figure}[!t]
\centering
\subfloat[]{
\includegraphics[width=2.55in]{ 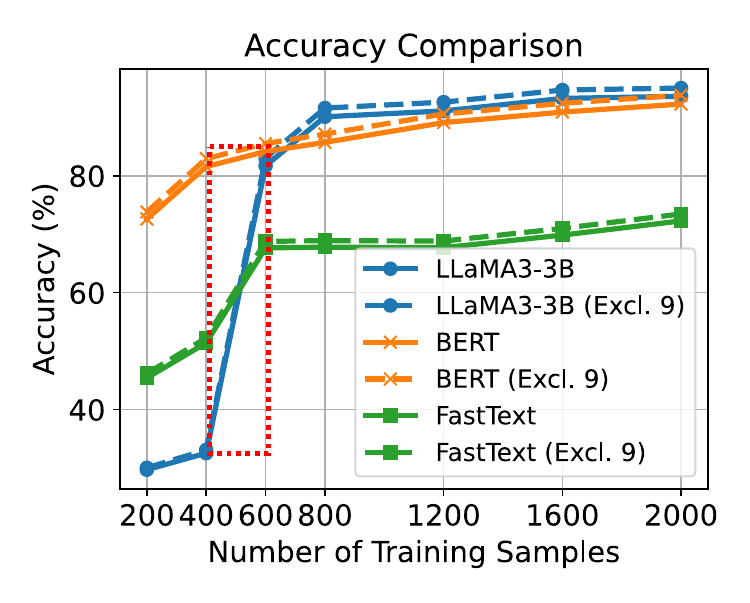}
\label{fig:image1}}
\subfloat[]{\includegraphics[width=2.55in]{ 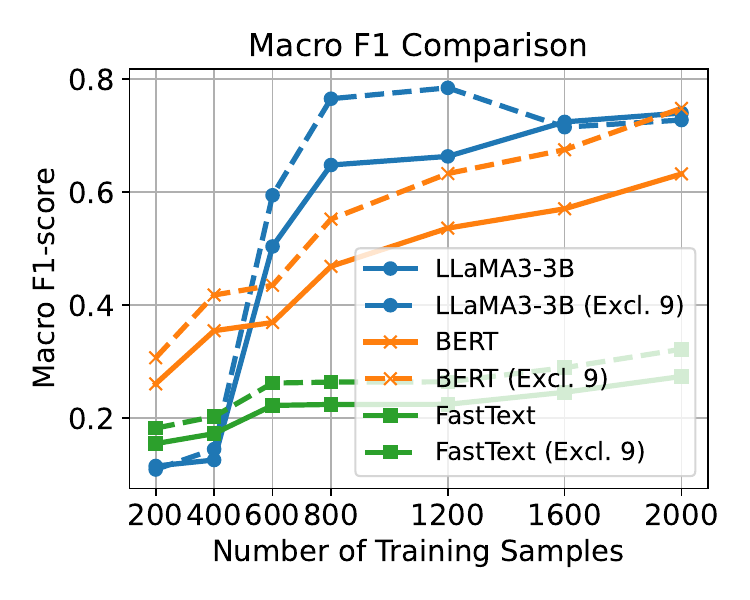}
\label{fig:image2}}
\caption{Accuracy (a) and Macro F1 (b) comparisons of LLaMA3-3B, BERT, and FastText under different numbers of training samples.}
\label{fig:number-sample}
\end{figure}

In this experiment, we fine-tune the models using training sets of varying sizes, ranging from 200 to 2,000 cases\footnote{BERT and FastText function primarily as feature extractors and require at least some labeled data for task-specific fine-tuning to perform classification effectively. Without labeled data, they can only provide general-purpose text representations and are unable to produce meaningful classification results.}. As shown in Figure~\ref{fig:number-sample}, with fewer than 400 examples, LLaMA3-3B requires more data to converge due to its longer prompts and bigger model size, whereas BERT and FastText, with simpler architectures and shorter inputs, achieve better performance on smaller datasets. However, once the training set reaches around 500 samples (the red box in Figure~\ref{fig:number-sample}a), LLaMA3-3B experiences an “aha stage” with performance rising sharply and surpassing both BERT and FastText. Although LLM is more data-hungry at the beginning, it can learn from both prompt and training data more effectively once sufficient training data is available, leading to superior accuracy and Macro F1 performance.

\subsubsection*{Consistency analysis}

\begin{table}[ht] 
\centering 
\caption{Self-Model and Cross-Model Consistency Table}
\scriptsize
\setlength{\tabcolsep}{3pt} 
\renewcommand{\arraystretch}{1} 
\begin{tabular}{l|ccccccccc}
\toprule
\multicolumn{5}{l}{\textbf{Overall consistency}}& \\
\hline
\multicolumn{3}{l}{}&\multicolumn{3}{c}{Models only} & \multicolumn{3}{c}{Including ground truth (GT)}\\
\hline
\multicolumn{3}{l}{Including: \textit{9 - Unknown}}&\multicolumn{3}{c}{ 0.9443} & \multicolumn{3}{c}{ 0.9438}\\
\multicolumn{3}{l}{Excluding: \textit{9 - Unknown}}&\multicolumn{3}{c}{0.9502}&\multicolumn{3}{c}{ 0.9508}\\
\midrule
\multicolumn{7}{l}{\textbf{Self-Model and Cross-Model consistency including: \textit{9 - Unknown}}}& \\
\hline
&BERT& LLaMA3 & LLaMA3 & LLaMA3 & Qwen2.5 & Mistral & LLaMA3 & GPT & GT \\ 
&& 1B & 3B & 8B & 7B & 7B & 70B & 4o &  \\ 
\hline
BERT &{1.00 }&{0.87 } & {0.94 } & { 0.94} & {0.93 } & {0.89} & { 0.91} & { 0.90} & {0.93 } \\ 
LLaMA3-1B & {0.87 }&0.96 & {0.86} & {0.86} & {0.86} & {0.83} & {0.84} & {0.84} & {0.86} \\ 
LLaMA3-3B & {0.94 }&{0.86} & {0.98} & {0.97} & {0.95} & {0.91} & {0.93} & {0.91} & {0.95} \\ 
LLaMA3-8B &{0.94} &{0.86} & {0.97} & 1.00 & 0.96 & {0.92} & {0.93} & {0.92} & 0.96 \\ 
Qwen2.5-7B & {0.93 }&{0.86} & {0.95} & {0.96} & {0.99} & {0.91} & {0.92} & {0.92} & {0.94} \\ 
Mistral 7B &{0.89} &{0.83} & {0.91} & {0.92} & {0.91} & {0.98} & {0.89} & {0.90} & {0.91} \\ 
LLaMA3-70B & {0.91}&{0.84} & {0.93} & {0.93} & {0.92} & {0.89} & {0.98} & {0.92} & {0.91} \\ 
GPT-4o & {0.90} &{0.84} & {0.91} & {0.92} & {0.92} & {0.90} & {0.92} & {0.97} & {0.91} \\ 
GT & {0.93} &{0.86} & {0.95} & {0.96} & {0.94} & {0.91} & {0.91} & {0.91} & {1.00} \\
\bottomrule
\end{tabular}
\label{tab:mancoll-cons}
\end{table}

To evaluate the stability of the models, we designed a set of consistency experiments, which include self-consistency, measuring the stability of repeated outputs from the same model, and cross-model consistency, measuring the agreement between outputs produced by different models. Since the accuracy of machine learning based methods is considerably lower than that of transformer-based models, we report consistency results only for transformer-based approaches.

Consistency analysis results are summarized in Table~\ref{tab:mancoll-cons}. Overall, the majority of models show high levels of both self-consistency and cross-model consistency. Besides, the overall consistency results (upper part of table) reveal that excluding ground truth (GT) label slightly increases consistency scores, and removing the \textit{Unknown} label further improves them. This suggests that some inconsistencies may stem from labeling errors: for example, instances marked as \textit{Unknown} in the dataset were consistently assigned to a specific class by the fine-tuned models, and these models strongly agreed on such classifications. This indicates that the models may have identified correct answers that were mislabeled in the ground truth.

In terms of self-consistency, all models achieve scores above 0.96, indicating that their outputs remain stable across repeated runs. Regarding cross-model consistency, the majority of model pairs achieve scores exceeding 0.90, highlighting strong agreement between different fine-tuned models. When comparing against the GT, fine-tuned models such as BERT, LLaMA3-3B, LLaMA3-8B, and Qwen2.5-7B achieve significantly higher consistency than non-fine-tuned models like GPT-4o and LLaMA3-70B. This trend aligns with the improvements previously reported in Table~\ref{Main-table-of-MANCOLL}.   

\subsection{Crash type}

Compared with \textit{Manner of Collision}, the extraction of \textit{Crash Type} is more challenging. There are 98 categories in total, and due to the hierarchical structure, shown in Figure~\ref{fig:crash-type-conf-cat}, the set of candidate \texttt{CRASHTYPE} is determined by the associated \texttt{CRASHCONF}. This decomposes the task into 13 smaller classification subtasks (one for each \texttt{CRASHCONF}), with the largest subtask containing up to 14 classes. Another source of difficulty is that a single crash may involve multiple vehicles. When classifying the \texttt{CRASHTYPE} for one vehicle, the narrative often contains descriptions of other vehicles, which can interfere with or complicate the judgment for the target vehicle.

Considering the above, it remains challenging for LLMs to extract information even though all tasks are within the traffic safety domain. To better understand how model performance can be improved when encountering such complex task, we further experiment with different LoRA projection configurations and examine their impact on fine-tuning results.
As described earlier in Section~\ref{sec:dataset}, \texttt{CRASHCONF} classification is a relatively easy task to achieve high accuracy. Therefore, in this part we treat \texttt{CRASHCONF} as oracle knowledge and focus only on the classification of \texttt{CRASHTYPE}. 

\subsubsection*{Results of different LLM on \textit{Crash Type}}
One crash may involve multiple vehicles and other vehicle descriptions can affect the prediction for the target vehicle. To account for this factor, we evaluate model performance by analyzing classification accuracy under different vehicle-count settings, grouping crashes into four categories depending on how many vehicles were involved in the crash: 1, 2, 3, and more than 3 vehicles. Given the difficulty of this task, we restricted the evaluation to the models that showed strong performance in the \textit{Manner of Collision} classification task, namely BERT and LLMs.

Table~\ref{Main table of Crashtype} reports the performance of different models on \texttt{CRASHTYPE} classification under varying numbers of vehicles per crash. The results show that this is a highly challenging task: without fine-tuning, most models achieve very low accuracy, typically in the range of 1\%$\sim$25\%. However, after parameter-efficient fine-tuning, all models see substantial improvements. In particular, LLaMA3-3B, LLaMA3-8B, and Qwen2.5-7B consistently reach close to $\sim80\%$ accuracy across different vehicle settings. Even the smallest 1B model, once adapted to the traffic safety domain, can surpass GPT-4o and the much larger LLaMA3-70B. Moreover, these smaller models achieve such performance with significantly lower training costs (less than 1 GPU-hour) and inference requirements, striking an effective balance between efficiency and accuracy. 
\begin{table}[t!]
\scriptsize
\centering
\caption{Performance of different LLMs for \texttt{CRASHTYPE} classification under different settings. The best results are highlighted in bold.}
\setlength{\tabcolsep}{10pt} 
\renewcommand{\arraystretch}{0.8} 
\begin{tabular}{lccrrrr}
\toprule
\multirow{2}{*}{\textbf{Backbones}} 
& \multirow{2}{*}{\textbf{Training step}} 
& \multirow{2}{*}{\makecell{\textbf{Training time}\\\textbf{(s)}}} 
& \multicolumn{4}{c}{\textbf{Number of vehicles in a crash}} \\
\cmidrule(l){4-7}
 & & & \textbf{=1} & \textbf{=2} & \textbf{=3} & \textbf{$>$3} \\
\midrule
\multicolumn{7}{l}{\textbf{Open source models}} \\
\midrule
\multirow{3}{*}{BERT} 

& 876 & 126.4&45.8 & 42.0&63.8 &72.1 \\
& 1460 & 252.9& 57.6 & 53.8 & 68.6 & 78.2\\
& 2920  & 506.6 &\textbf{68.9} &\textbf{68.8}& \textbf{77.1}& \textbf{78.2}\\
\midrule
\multirow{5}{*}{LLaMA3-1B} 
& \multicolumn{2}{c}{Original} &7.1 & 3.6&3.7 &1.3 \\
\cmidrule(l){2-7}
& 721  & 198.4 &49.1 &26.7& 43.3& 45.1\\
& 1442 & 397.2&43.7 & 42.7&64.1 &75.2 \\
& 2163 & 594.7 & 52.6&49.4 &64.5 & 73.4\\
& 2884 & 780.3& \textbf{62.2} & \textbf{51.4} &\textbf{64.7} & \textbf{75.4}\\
\midrule
\multirow{5}{*}{LLaMA3-3B} 
& \multicolumn{2}{c}{Original} & 50.2 & 23.3 & 18.4 & 12.4 \\
\cmidrule(l){2-7}
& 721  & 430.1 &  73.9 & 53.7 & 67.1 & 77.4 \\
& 1442 & 842,9 &  74.0 & 69.5 & 78.9 & 81.9 \\
& 2163 & 1249.8 &  73.6 & 71.5 & 82.6 & 83.9 \\
& 2884 & 1683.7 &  \textbf{76.0} & \textbf{73.7} & \textbf{81.8} & \textbf{83.6} \\
\midrule
\multirow{5}{*}{Qwen2.5-7B} 
& \multicolumn{2}{c}{Original} & 25.3 & 21.7 & 14.8 & 6.2 \\
\cmidrule(l){2-7}
& 721  & 659.3  & 77.2 & 57.6 & 68.9 & 70.3 \\
& 1442 & 1308.5 & 78.3 & 68.9 & 72.8 & 79.4 \\
& 2163 &  1965.2 & \textbf{79.1} & \textbf{70.3} & \textbf{80.6} & \textbf{80.9} \\
& 2884 & 2605.5 &77.2&72.9 & 80.1& 81.5\\
\midrule
\multirow{5}{*}{LLaMA3-8B} 
& \multicolumn{2}{c}{Original}  & 40.4 & 18.9 & 19.0 & 15.1 \\
\cmidrule(l){2-7}
& 721  & 694.3  & 73.9 & 53.7 & 67.1 & 75.2 \\
& 1442 & 1387.5 & 76.6 & 77.0 & 83.3 & 83.6 \\
& 2163 & 2088.5  & \textbf{77.1} & \textbf{77.3} & \textbf{82.0} & \textbf{84.8} \\
& 2884 & 2780.7  & 77.6 & 77.3 & 81.9 & 81.2\\
\midrule
\multirow{1}{*}{LLaMA3-70B} 
& \multicolumn{2}{c}{Original} &72.7 &41.8&44.3& 55.0\\
\midrule
\multicolumn{7}{l}{\textbf{Closed models}} \\
\midrule
\multicolumn{3}{l}{GPT-4o} &  45.3 & 64.3 & 58.8 & 70.3 \\
\bottomrule
\end{tabular}

\label{Main table of Crashtype}

\end{table}

When analyzing results across different vehicle counts, we observe that collisions involving exactly two vehicles yield the lowest accuracy for all models. This is to be expected. Single-vehicle cases are the easiest, as the crash narrative only describes one vehicle and contain no textual interference from other vehicles. Multi-vehicle crashes involving three or more vehicles often correspond to chain collisions (e.g., multi-vehicle rear-end crashes), where multiple vehicles share the same \textit{Crash Type}. This homogeneity reduces ambiguity and makes classification more straightforward, despite the larger number of vehicles. The most challenging setting resides in crashes involving exactly two vehicles. In these cases, the narrative usually mixes descriptions of both vehicles, and their \textit{Crash Types} are often different. Consequently, two-vehicle crashes represent the most difficult scenario, demanding stronger reasoning and disambiguation capabilities from the models.

\begin{figure}[!t]
\centering
\subfloat[Fine-tuned on query projection.]{
\includegraphics[width=1.5in, trim=0 0 90 0, clip]{ 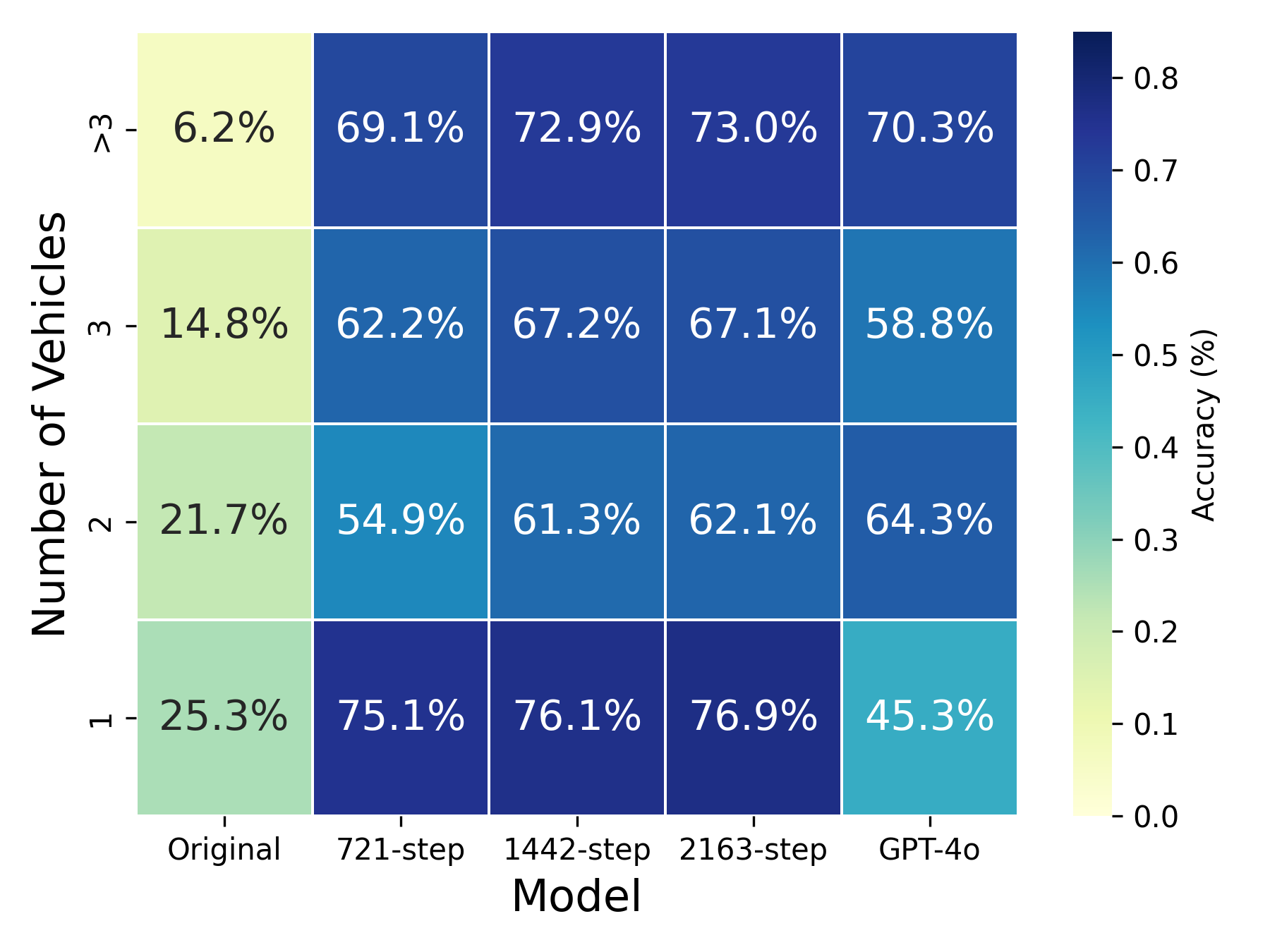}
\label{fig:image1}}
\hfil
\subfloat[Fine-tuned on query and value projection.]{\includegraphics[width=1.5in, trim=0 0 90 0, clip]{ 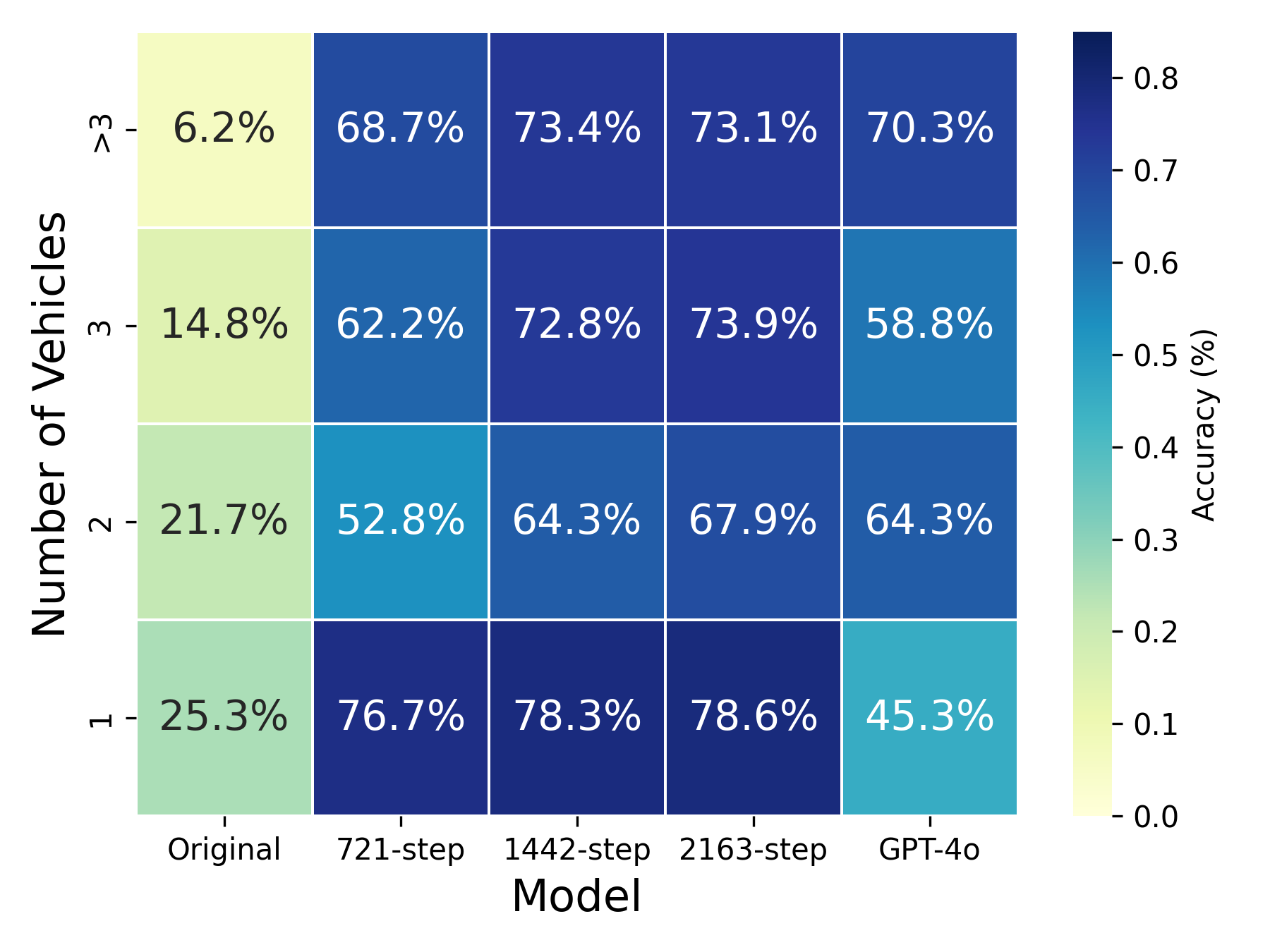}
\label{fig:image1}}
\hfil
\subfloat[Fine-tuned on query, key and value projection.]{\includegraphics[width=1.9in, trim=0 0 0 0, clip]{ 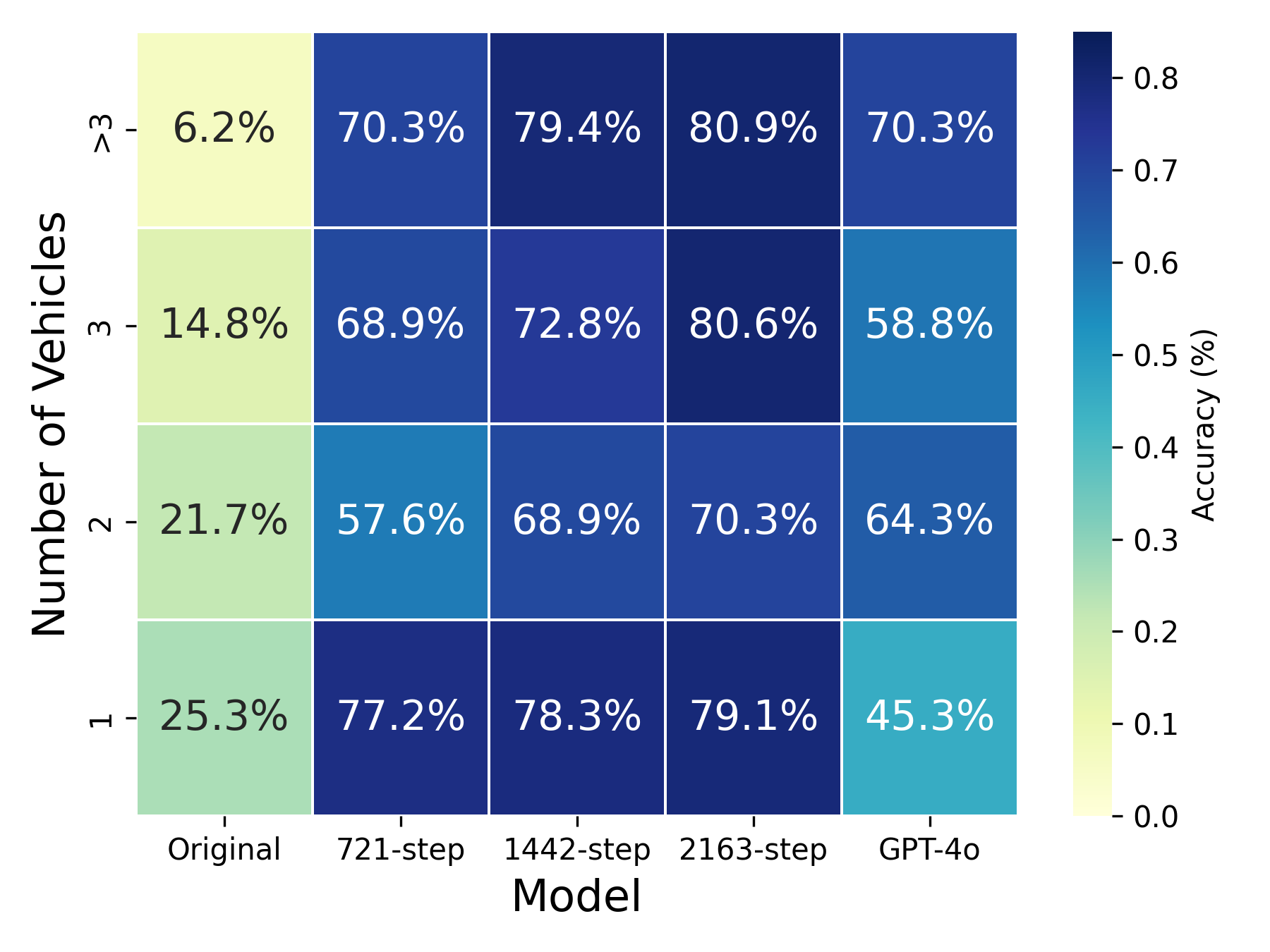}
\label{fig:image2}}
\caption{Accuracy (\%) across different numbers of vehicles for various fine-tuning strategies. (a) Fine-tuning only the query projection, (b) fine-tuning both query and value projections, and (c) fine-tuning query, key, and value projections. }
\label{fig:ablation2}
\end{figure}

\subsubsection*{Effect of LoRA Training Strategies}
Given the complexity of the \textit{Crash Type} classification task, we further explore how LoRA-adapted parameters impacts performance. Specifically, we vary which attention projection matrices to be trained under LoRA, selecting from the query, key, and value projections $W_Q, W_K, W_V \in \mathbb{R}^{d \times k}$. We consider three configurations: (i) LoRA on $W_Q$ only; (ii) LoRA on $W_Q$ and $W_V$; and (iii) LoRA on all three, $W_Q$, $W_K$, and $W_V$\footnote{Since the self-attention output is a weighted sum of values, fine-tuning different single projection matrices introduces additional low-rank weight on the same attention mechanism. Therefore, varying which of $W_Q$, $W_K$, and $W_V$ are fine-tuned does not drastically have a large impact, although increasing the number of projections equipped with LoRA adapters does.}. 

Results in Figure \ref{fig:ablation2} show that the accuracy increases monotonically with the number of adapted attention matrices: Updating $W_Q$ alone yields the weakest performance; adapting both $W_Q$ and $W_V$ performs better; and jointly adapting $W_Q$, $W_K$, and $W_V$ achieves the best results. We attribute this to the task’s requirement for token interactions: restricting adaptation to queries constrains the model’s ability to reshape attention patterns; co-adapting keys improves query–key alignment for retrieval, while adapting values allows the model to better propagate matched evidence through the representation. In short, enabling LoRA on multiple attention projections provides the necessary degrees of freedom to fit the \textit{Crash Type} task more effectively.

\subsubsection*{Consistency analysis}

Looking at the available statistics, crashes involving one single vehicle or two vehicles account for nearly the same amount of crashes, and together they represent over 93\% of all available crashes. Since such crashes are also the primary interest of researchers, our consistency analysis focuses on these two settings.

\begin{figure}[!t]
\centering

\subfloat[Single-vehicle setting]{
\includegraphics[width=2.55in, trim=5 0 5 0, clip]{ 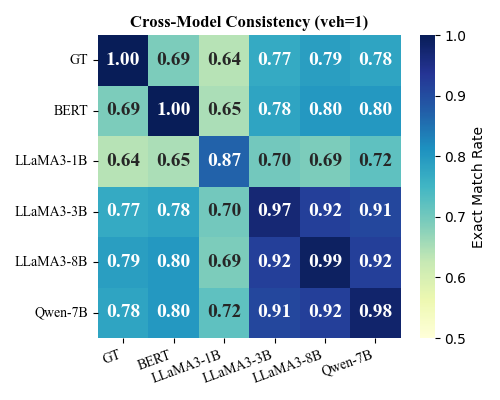}
\label{fig:cons-image1}}
\hfil
\subfloat[Two-vehicle setting]{
\includegraphics[width=2.55in, trim=5 0 5 0, clip]{ 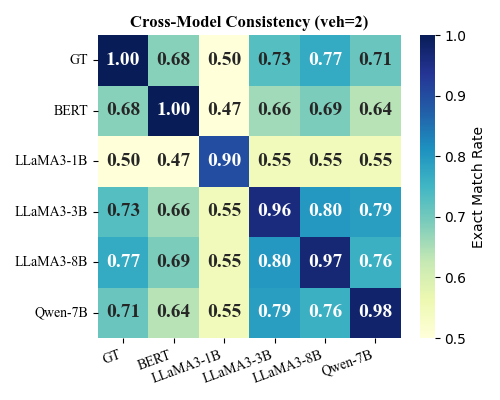}
\label{fig:cons-image2}}
\caption{Self-Model and Cross-Model Consistency}
\end{figure}

For single-vehicle setting, as shown in Figure~\ref{fig:cons-image1}, we observe that LLaMA3-3B, LLaMA3-8B and Qwen2.5-7B show near-perfect run-to-run agreement ($\sim 0.98$) on the diagonal, while smaller models LLaMA3-1B maintain a self-consistency ($\sim 0.87$). This indicates that fine-tuned models yield stable outputs across runs on \texttt{CRASHTYPE} classification. The agreements between LLaMA3-3B, LLaMA3-8B and Qwen2.5-7B are also high ($\sim 0.91$), suggesting that different LLMs converge to similar predictions despite variying in size and architecture. When compared with CISS ground-truth labels (GT), the average consistency is slightly lower ($\sim 0.78$), reflecting the possible presence of annotation noise and labeling errors in the dataset, which LLMs may help to mitigate.

For the two-vehicle setting, as shown in Figure~\ref{fig:cons-image2}, we observe a similar trend as in the single-vehicle case: larger models such as LLaMA3-3B, LLaMA3-8B, and Qwen2.5-7B maintain high self-consistency, while smaller models like LLaMA3-1B achieve a reasonable level of self-consistency ($\sim 0.87$). However, cross-model agreements among LLaMA3-3B, LLaMA3-8B, and Qwen2.5-7B drop noticeably (to around $0.79$). Compared with CISS ground-truth labels, the average cross-consistency ($\sim 0.78$) is also lower than in the single-vehicle setting. All above reflect a reduced accuracy because of the added complexity and ambiguity of multi-vehicle crash narratives.
\section{Data Analysis}

\subsection{Distribution of reclassified \textit{Unknown} \texttt{MANCOLL} cases}
\begin{figure}[!t]
\centering
\subfloat[]{
\includegraphics[width=3.5in, trim=55 0 00 0, clip]{ 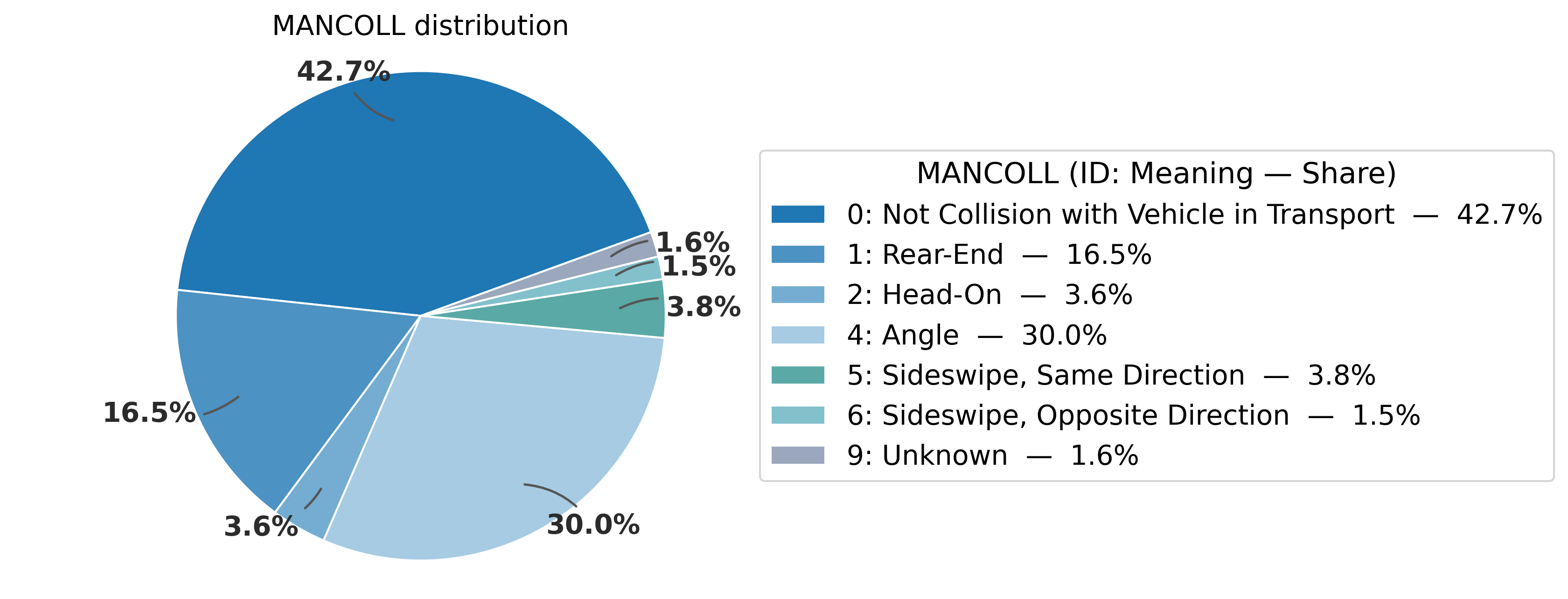}
\label{fig:image1}}
\hfil
\subfloat[]{\includegraphics[width=1.6in, trim=0 0 00 0, clip]{ 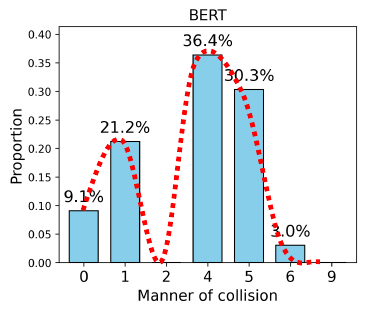}
\label{fig:image1}}
\hfil
\subfloat[]{\includegraphics[width=1.6in, trim=0 0 00 0, clip]{ 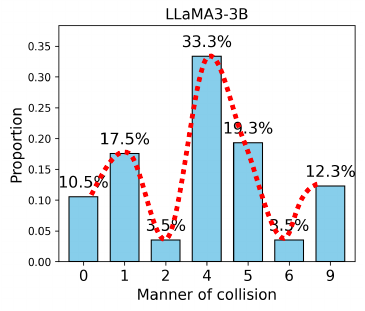}
\label{fig:image1}}
\hfil
\subfloat[]{\includegraphics[width=1.6in, trim=0 0 00 0, clip]{ 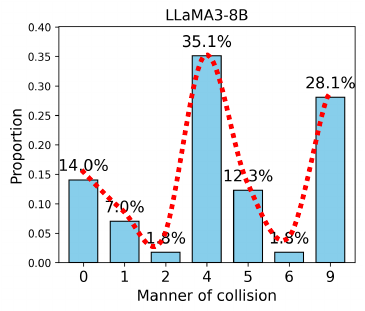}
\label{fig:image1}}
\hfil
\subfloat[]{\includegraphics[width=1.6in, trim=0 0 00 0, clip]{ 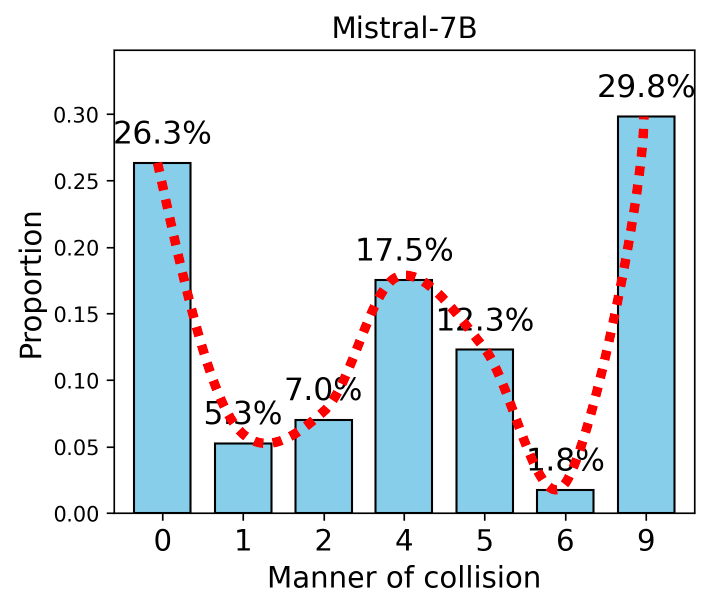}
\label{fig:image1}}
\hfil
\subfloat[]{\includegraphics[width=1.6in, trim=0 0 00 0, clip]{ 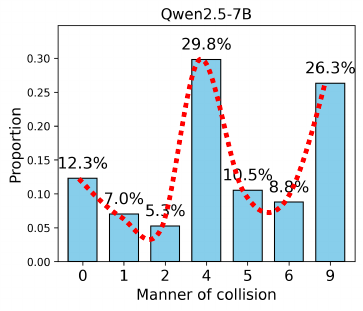}
\label{fig:image1}}
\hfil
\subfloat[]{\includegraphics[width=1.6in, trim=0 0 00 0, clip]{ 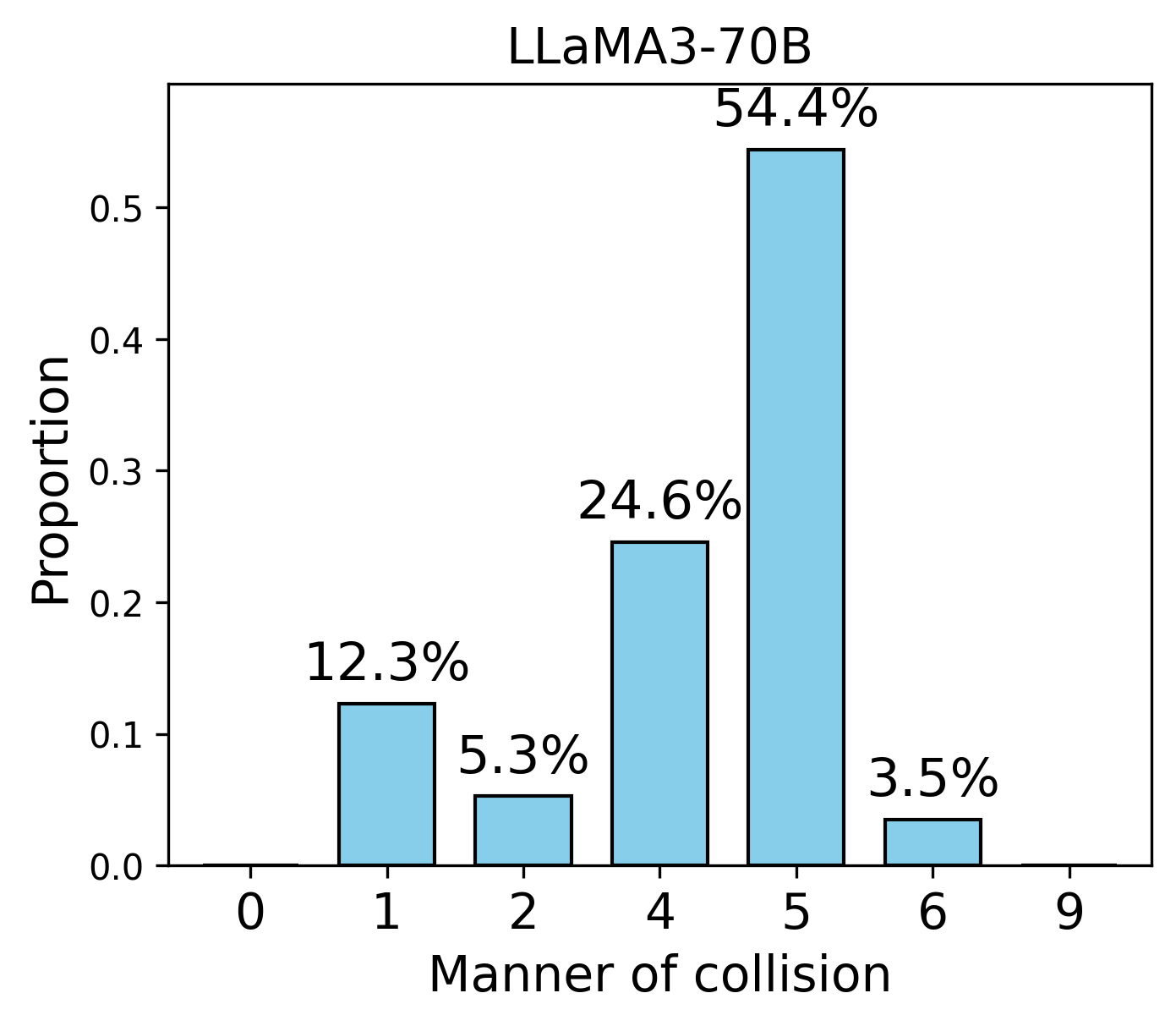}
\label{fig:image1}}
\hfil
\subfloat[]{\includegraphics[width=1.6in, trim=0 0 00 0, clip]{ 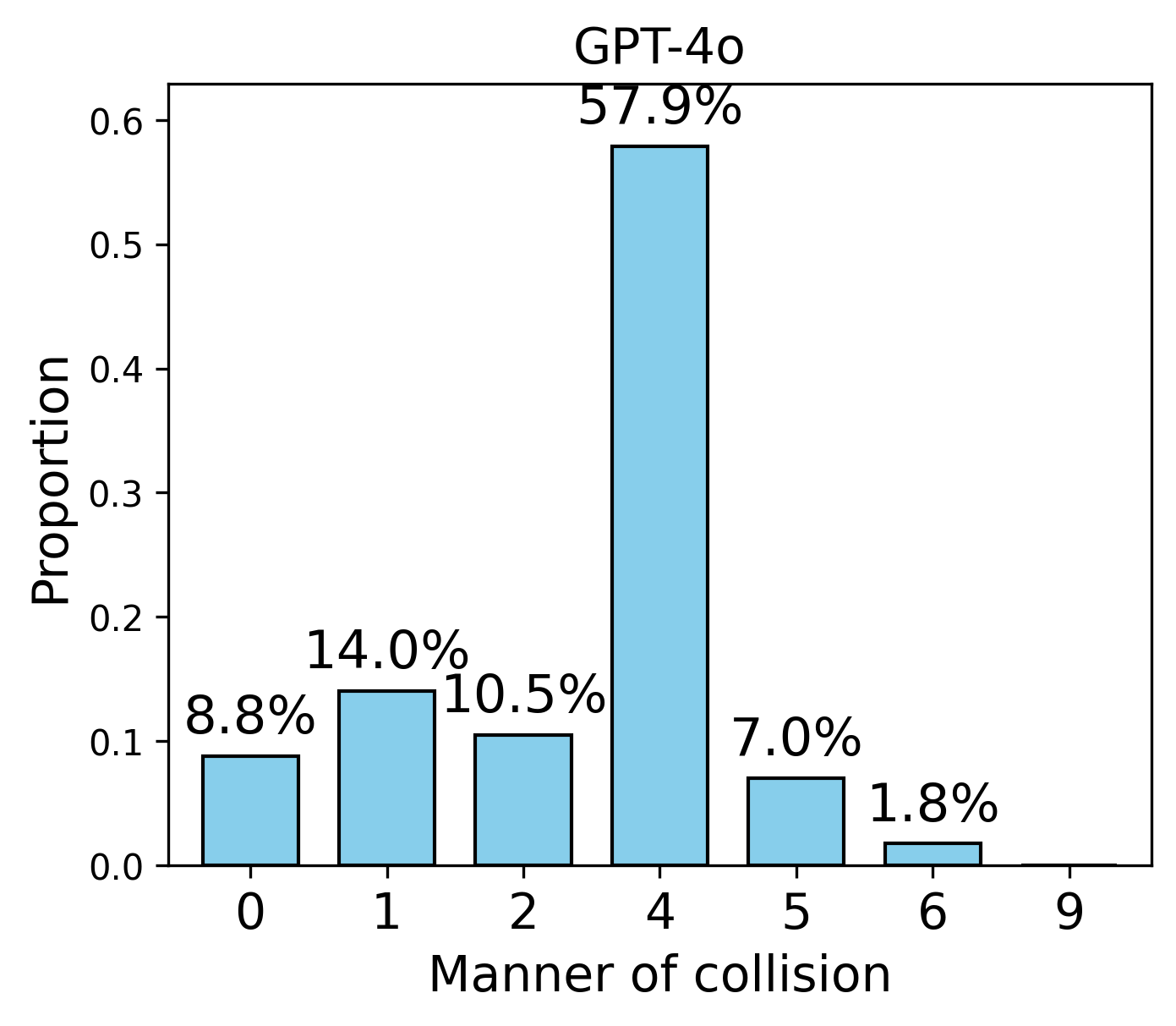}
\label{fig:image2}}
\caption{Distributional analysis of \texttt{MANCOLL} cases originally labeled as \textit{Unknown}. (a) The original distribution of \texttt{MANCOLL} in database, where \textit{Unknown} accounts for about 1.6\%. (b–h) LLM-based reclassification results across different models. The red dashed lines are added to smooth and highlight the overall distribution pattern.}
\label{fig:MANCOLL-class-unknown}
\end{figure}
In the original database, the proportion of samples labeled as \textit{Unknown} is about 1.6\%. Previous experiments demonstrated that LLMs achieve strong overall performance on the \texttt{MANCOLL} classification task, with some models reaching up to 97\% accuracy. Notably, the models were also able to reassign many of the samples originally labeled as \textit{Unknown} into more specific categories. To further investigate this behavior, we conducted a detailed analysis focusing on the predictions for the \textit{Unknown} subset.

As shown in Figure~\ref{fig:MANCOLL-class-unknown}, categories 2 (Head-On) and 6 (Sideswipe, Opposite Direction) consistently appear with very low proportions across all models. This is partly because these categories are rare in the original ground-truth annotations, and also because narrative descriptions often provide limited clues for such cases. In contrast, models (c)–(f) maintain a relatively stable distribution pattern, showing that PLMs not only resolve the ambiguity in \textit{Unknown} cases but also exhibit strong cross-model consistency, which enhances the reliability of the results.

\subsection{\texttt{CRASHTYPE} distributions: CISS annotations vs. LLM predictions}

To further verify that the labels generated by the LLM do not substantially alter the intrinsic characteristics of the data, we compared some statistics between LLM-generated labels and ground-truth. Here we primarily focus on accidents involving one or two vehicles, which together account for approximately 93\% of the dataset. For single-vehicle crashes, we conduct distributional analysis of \textit{Crash Type}, while for two-vehicle crashes, we perform correlation analysis to assess the association between the \textit{Crash Types}  of the two vehicles.
\subsubsection*{Single-vehicle setting }
To characterize single-vehicle crashes, we analyze the distribution of their \textit{Crash Types}. Specifically, we compute the frequency distribution of the most common crash categories (1-16), and the overall divergence of the distribution. Since some \textit{Crash Types} in the distribution may have zero or near-zero frequencies, we adopt Jensen–Shannon (JS) divergence~\citep{nielsen2019jensen} to measure the difference between the two distributions, as it is symmetric, bounded, and robust to zero-probability events, thereby avoiding the divergence issues inherent in Kullback–Leibler divergence~\citep{kullback1951kullback}. 

\begin{figure}[!t]
\centering
\subfloat[LLaMA3-1B]{
\includegraphics[width=1.65in, trim=0 0 0 20, clip]{ 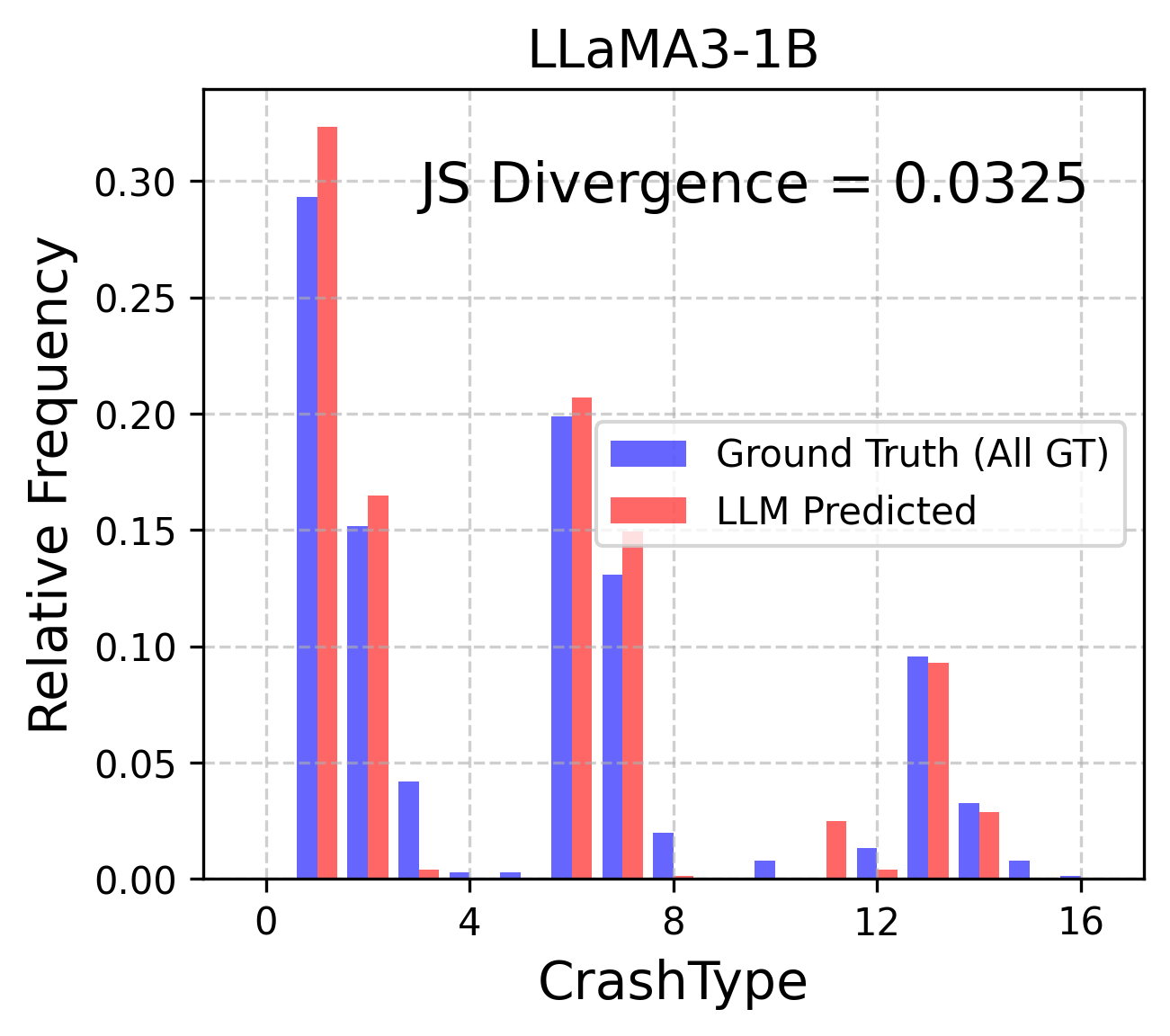}
\label{fig:image1}}
\hfil
\subfloat[LLaMA3-3B]{
\includegraphics[width=1.65in, trim=0 0 0 20, clip]{ 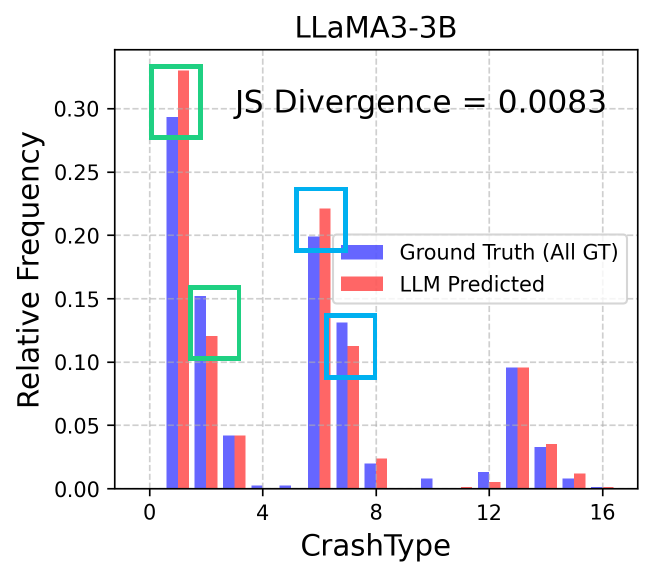}
\label{fig:image1}}
\hfil
\subfloat[LLaMA3-8B]{
\includegraphics[width=1.65in, trim=0 0 00 20, clip]{ 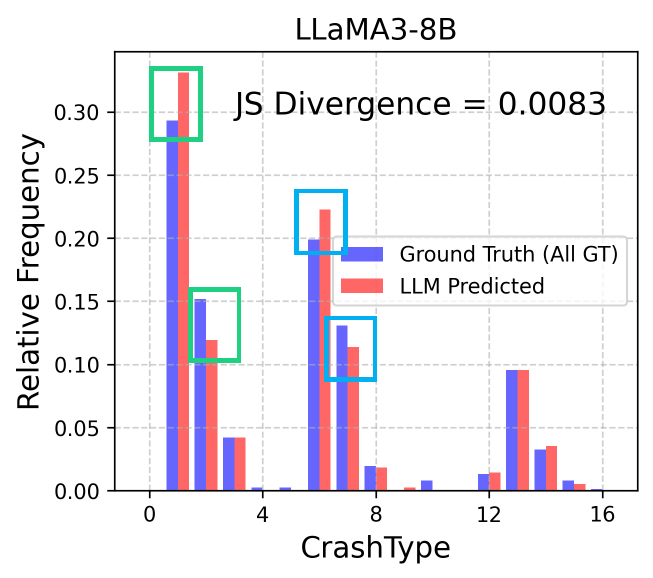}
\label{fig:image1}}
\hfil
\subfloat[Qwen2.5-7B]{
\includegraphics[width=1.65in, trim=0 0 00 20, clip]{ 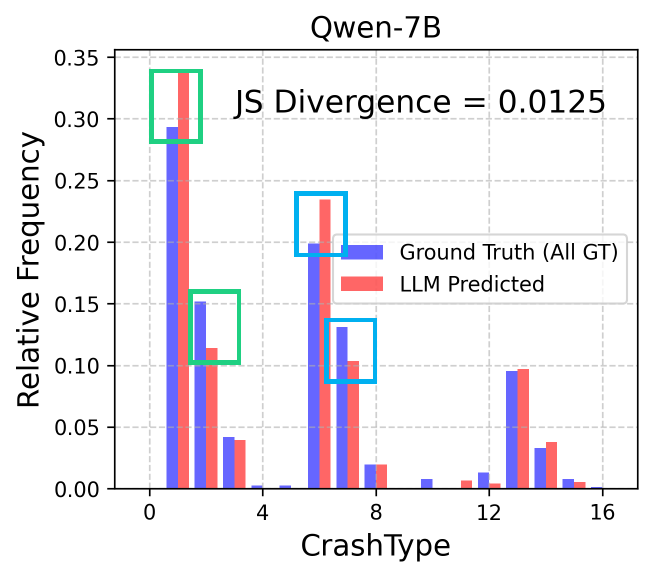}
\label{fig:image1}}
\hfil
\subfloat[LLaMA3-70B]{
\includegraphics[width=1.65in, trim=0 0 00 20, clip]{ 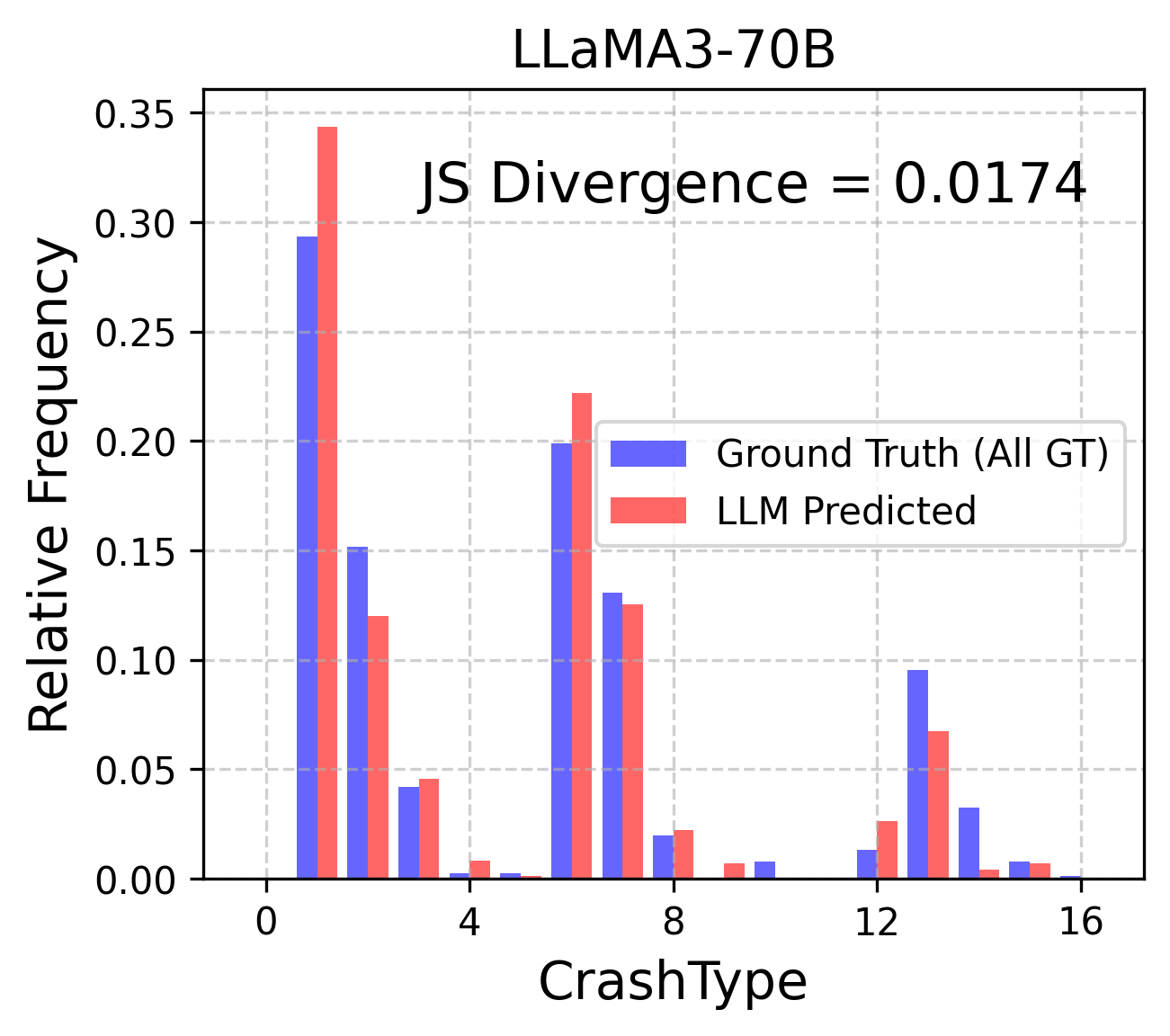}
\label{fig:image1}}
\hfil
\subfloat[GPT-4o]{
\includegraphics[width=1.65in, trim=0 0 00 20, clip]{ 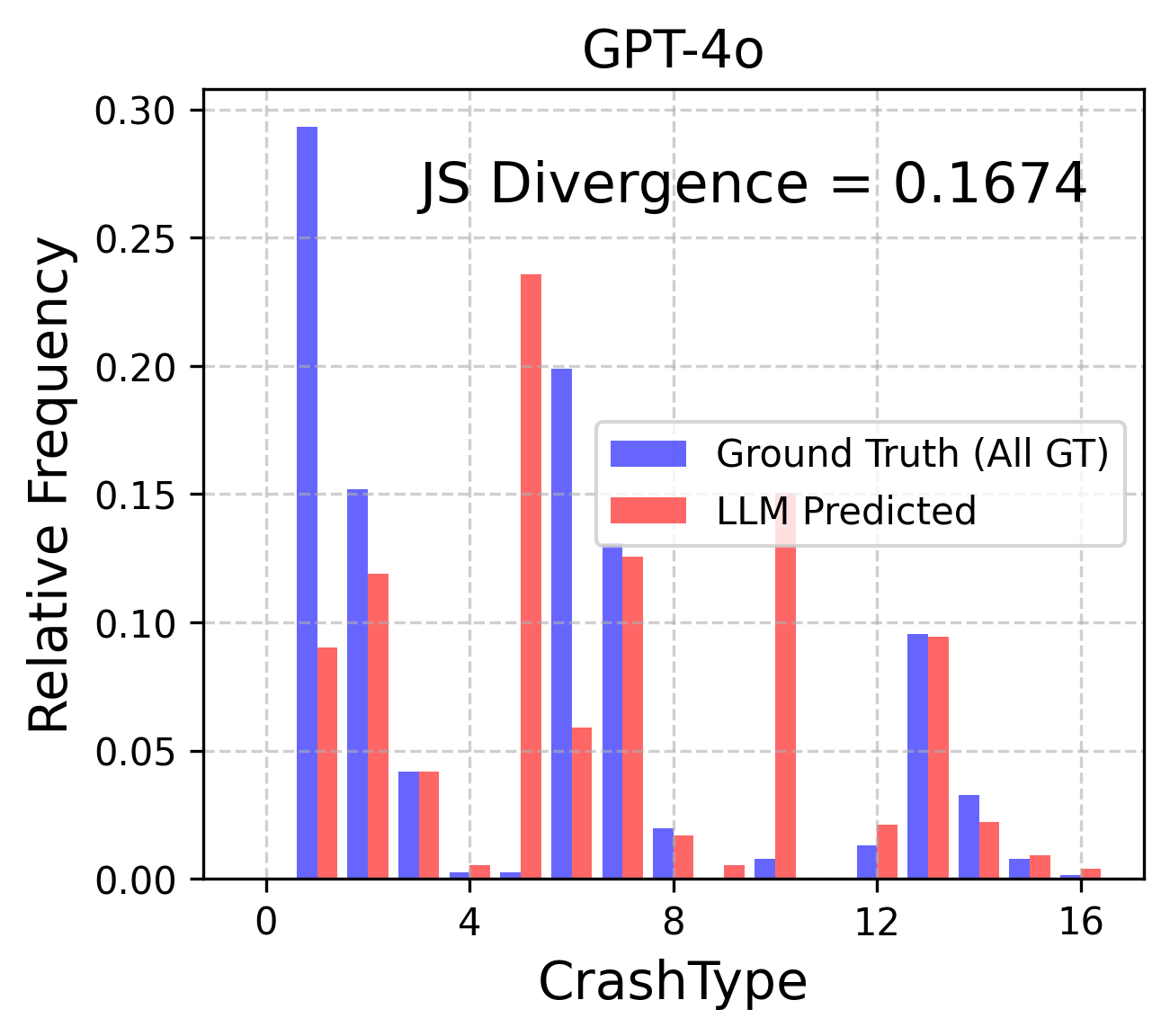}
\label{fig:image1}}
\hfil
\caption{Comparison of \texttt{CRASHTYPE} distributions between the CISS ground truth (blue) and LLM-predicted results (red) for crashes involving a single vehicle. Each subplot corresponds to a different model, with the JS divergence reported as a measure of distributional similarity.}
\label{fig:crashtype-bar}
\end{figure}

\begin{table}[h]
\footnotesize
\centering
\setlength{\tabcolsep}{3pt}
\caption{Examples of single-vehicle \texttt{CRASHTYPE} categories related to road departure}
\label{tab:crashtype-departure}
\begin{tabular}{clc}
\toprule
\textbf{Code} & \textbf{Description} & \textbf{Departure Direction} \\
\midrule
1 & Roadside departure under a controlled situation.  & Right \\
2 & Roadside departure because of lost traction or control. & Right \\
6 & Roadside departure under a controlled situation. & Left \\
7 & Roadside departure because of lost traction or control. & Left \\
\bottomrule
\end{tabular}
\end{table}

Figure~\ref{fig:crashtype-bar} presents the predicted distribution of single-vehicle \texttt{CRASHTYPE} across different models, where only the first 16 crash categories are included since single-vehicle data is restricted to this subset. Overall, the fine-tuned models successfully preserve the distributional characteristics of the ground-truth data: the relative frequency and ranking of each class are largely consistent with the ground truth labels. Moreover, except for the 1B model, all fine-tuned models achieve closer alignment with the ground-truth distribution compared to much larger non-fine-tuned models such as LLaMA3-70B and GPT-4o. Besides, noticeable discrepancies remain for certain categories, particularly \texttt{CRASHTYPE} 1, 2, 6, and 7\footnote{The detailed definitions of these \textit{Crash Types} are provided in Table~\ref{tab:crashtype-departure}}. The boxed regions in the histogram illustrate a complementary phenomenon between the ground truth distribution and the LLM-predicted results. Specifically, the discrepancies arise mainly from whether the model correctly captures the detail of losing control or not. Because the LLM fails to fully recognize this subtle distinction, its predicted frequencies are shifted compared to the ground truth. Interestingly, similar complementary patterns are consistently observed across fine-tuned models, which indicates that this misalignment is not accidental. Through manual inspection of the CISS dataset, we found that part of this mismatch arises from inconsistencies in the recorded ground truth in CISS, where some cases were not assigned correctly. Illustrative examples of such issues are provided in~\ref{app:ex-1veh}.

\subsubsection*{Two-vehicle setting}

\begin{figure}[!t]
\centering
\subfloat[LLaMA3-1B]{
\includegraphics[width=1.65in, trim=0 0 0 0, clip]{ 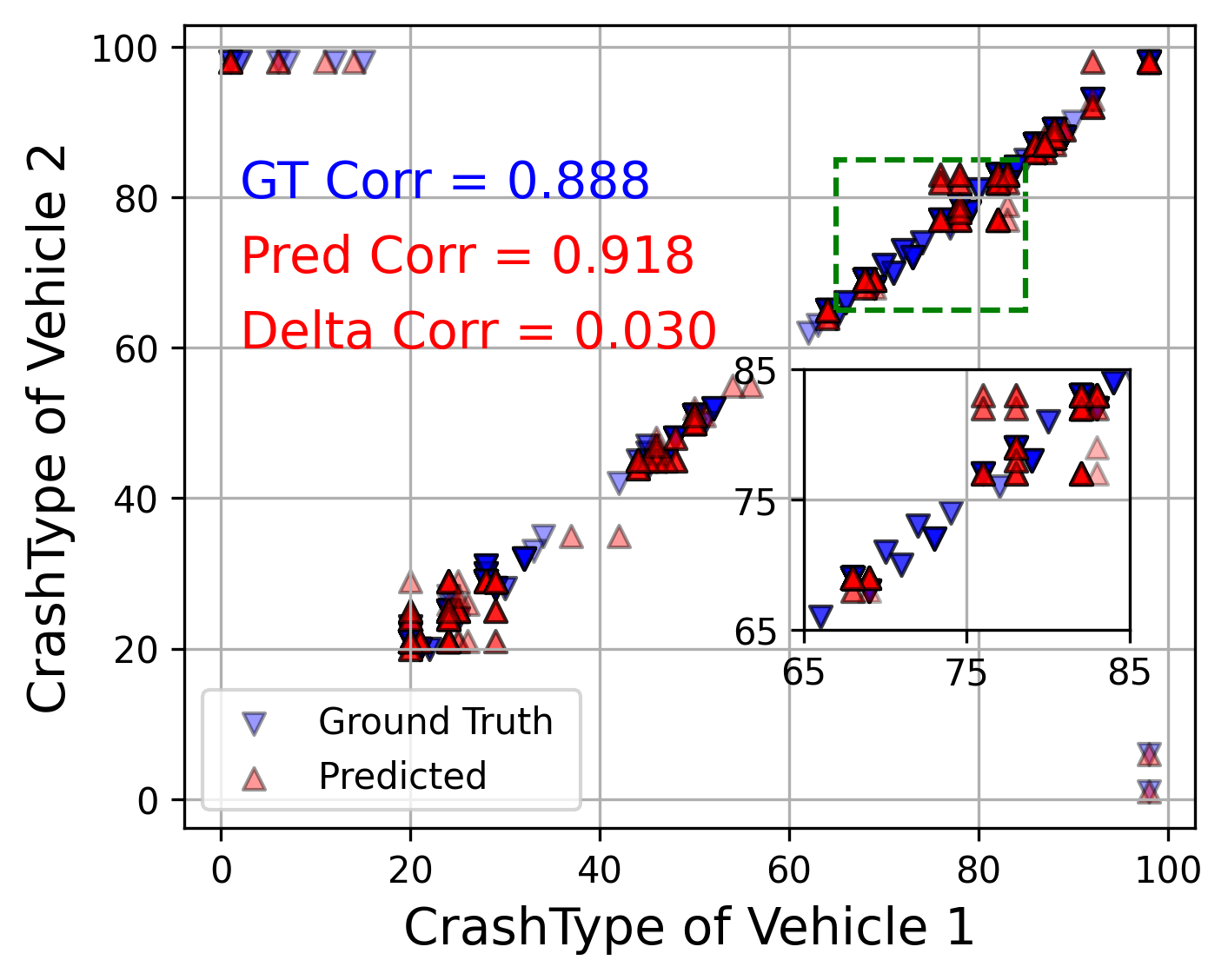}
\label{fig:image1}}
\hfil
\subfloat[LLaMA3-3B]{
\includegraphics[width=1.65in, trim=0 0 0 0, clip]{ 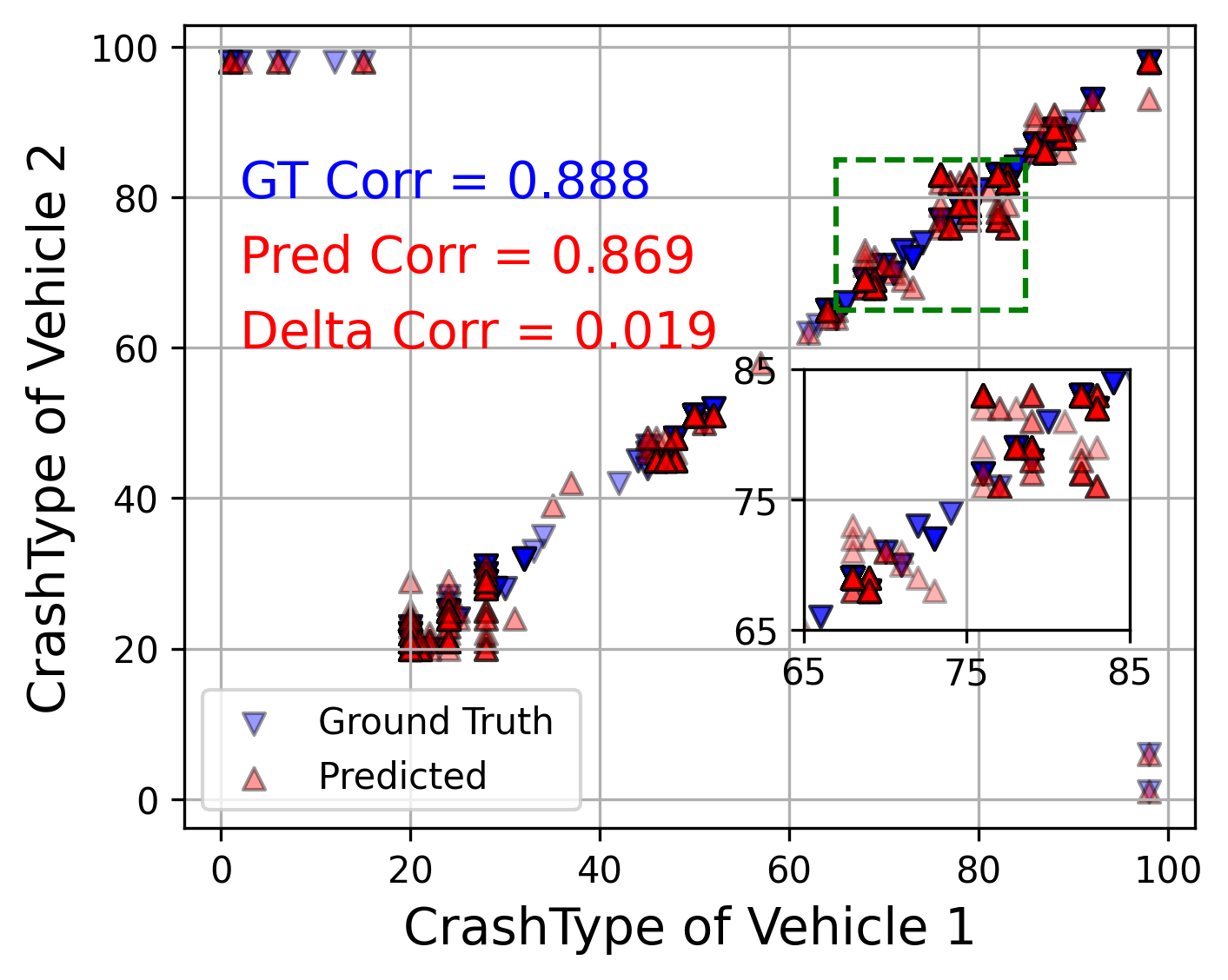}
\label{fig:image1}}
\hfil
\subfloat[LLaMA3-8B]{
\includegraphics[width=1.65in, trim=0 0 00 0, clip]{ 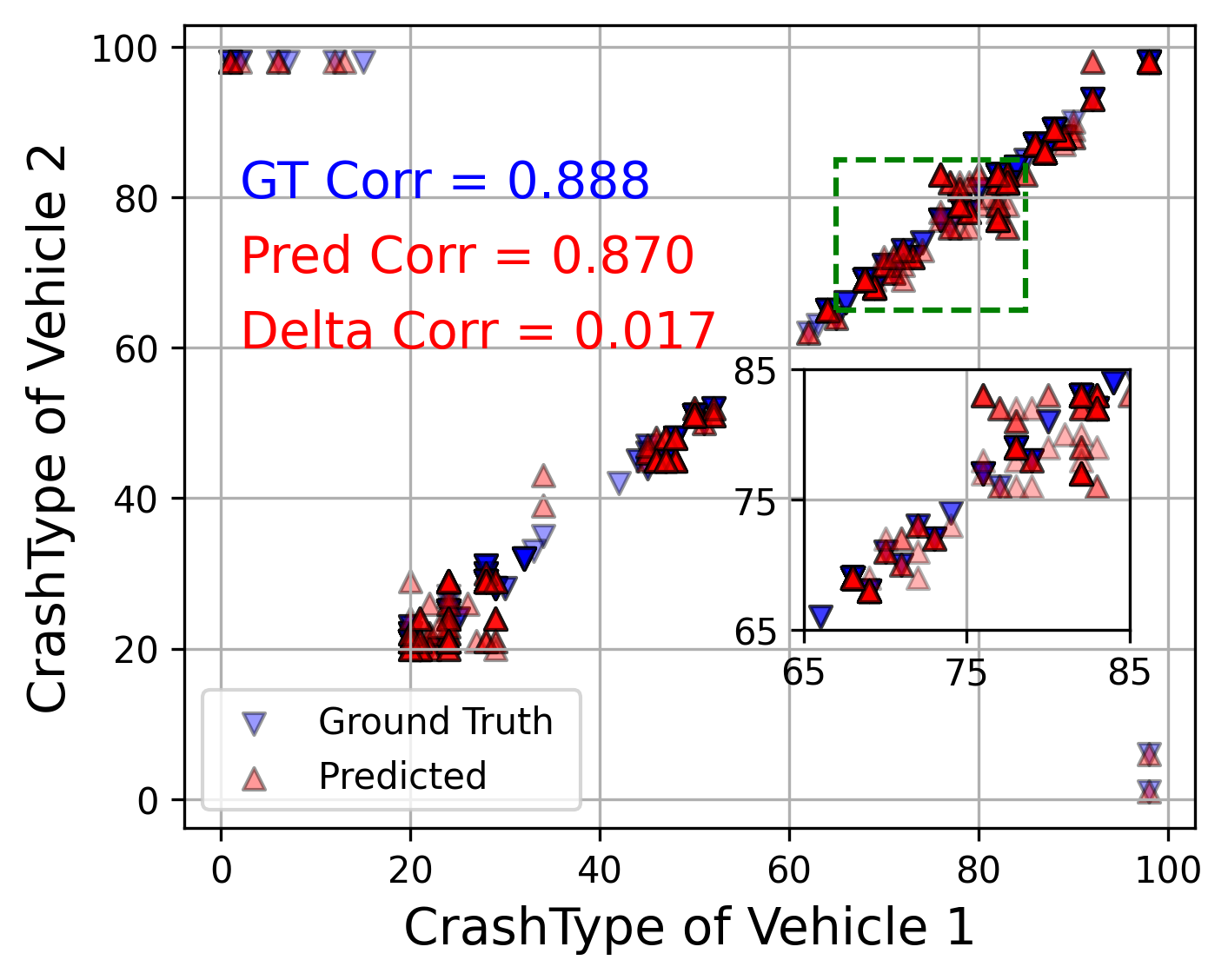}
\label{fig:image1}}
\hfil
\subfloat[Qwen2.5-7B]{
\includegraphics[width=1.65in, trim=0 0 00 0, clip]{ 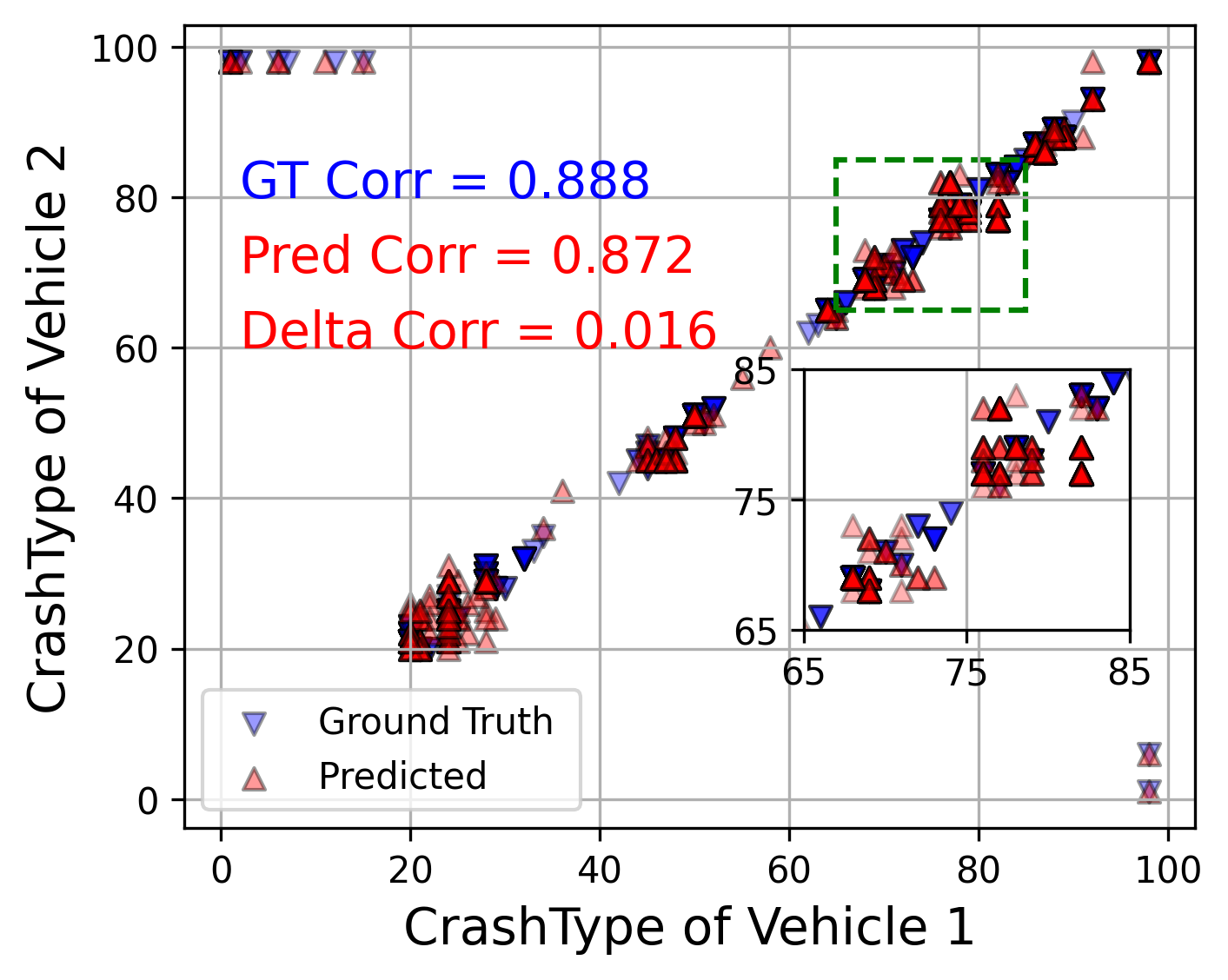}
\label{fig:image1}}
\hfil
\subfloat[LLaMA3-70B]{
\includegraphics[width=1.65in, trim=0 0 00 0, clip]{ 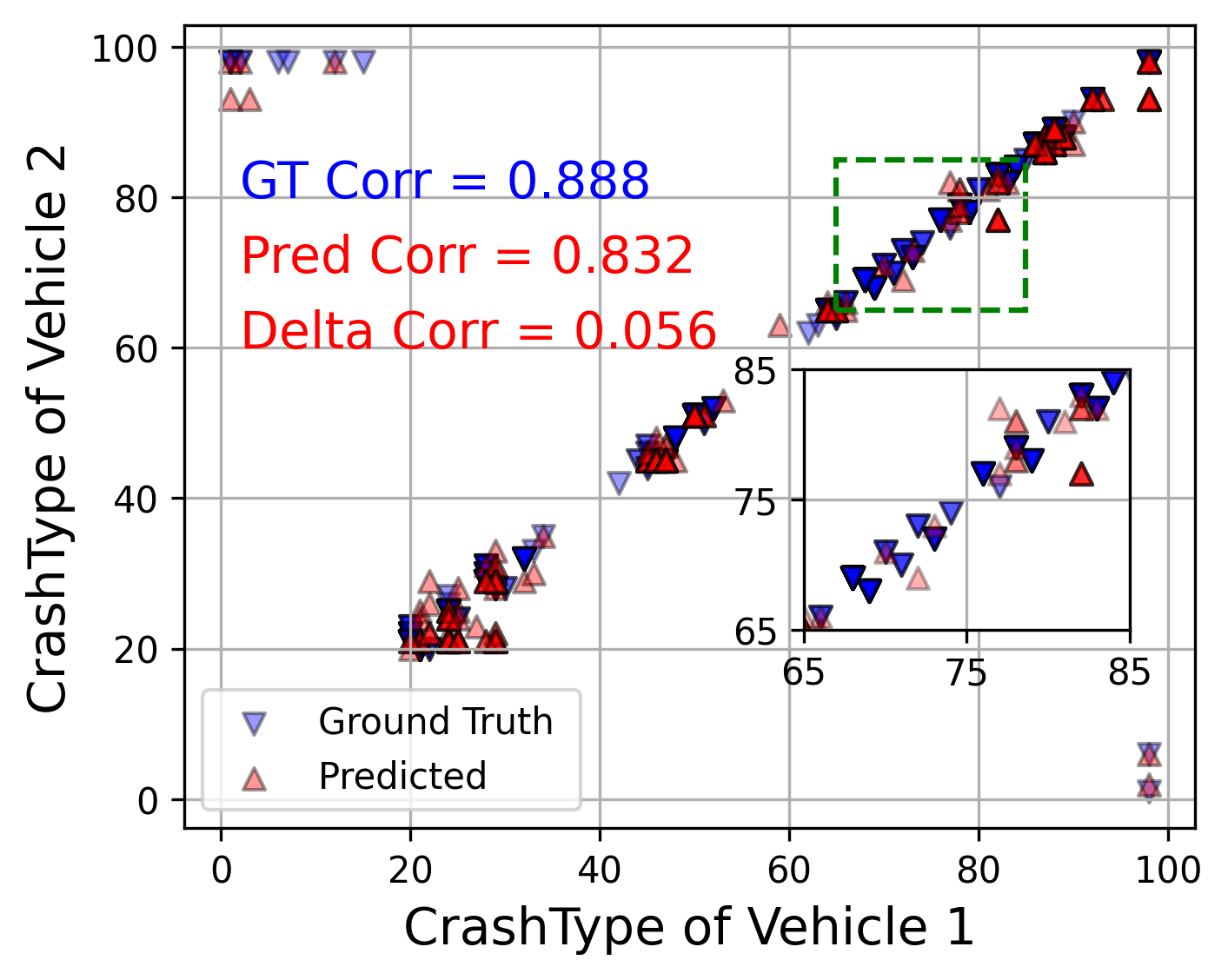}
\label{fig:image1}}
\hfil
\subfloat[GPT-4o]{
\includegraphics[width=1.65in, trim=0 0 00 0, clip]{ 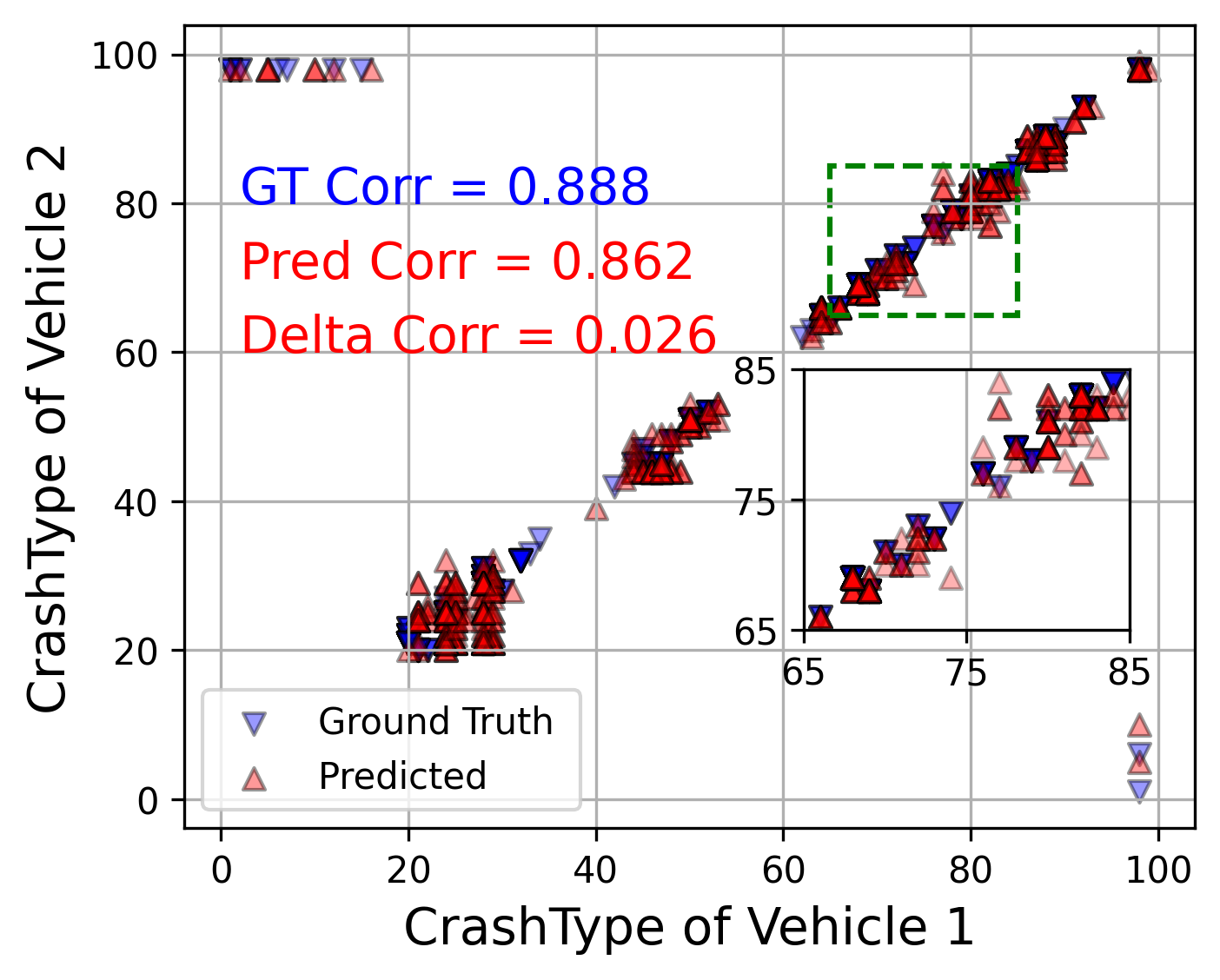}
\label{fig:image1}}
\hfil
\caption{Correlation analysis of \texttt{CRASHTYPE} combinations between two vehicles in the same crash. Blue markers represent CISS ground truth (GT) and red markers represent LLM predictions, with Spearman correlation coefficients reported for both. }
\label{fig:crashtype-combinations}
\end{figure}

Crash types of vehicles in the same accident are naturally related. To quantify this, we compute correlation coefficients using Kendall’s Tau~\citep{sen1968estimates}. Since the \textit{Crash Type} values are discrete identifiers rather than continuous quantities, \textit{Kendall’s Tau} is more appropriate than \textit{Pearson} or \textit{Spearman} correlation~\citep{arndt1999correlating}. It is important to note that in this task, a higher correlation is not necessarily better; instead, what matters is whether the correlation obtained from the predicted labels is closely aligned with the ground-truth correlation.

As shown in Figure~\ref{fig:crashtype-combinations}, models with higher classification accuracy generally yield correlations closer to ground truth. Across all fine-tuned models, the predicted correlations remain close to the true values, indicating that they not only achieve accurate \textit{Crash Type} predictions but also preserve the inter-vehicle dependency patterns in the CISS dataset.

\section{Discussion}
In this section we discuss the results with respect to task-specific semantics, data annotation, quality and quantity of the data and model performance. 

\textbf{Task-specific semantics}. We investigate whether fine-tuned LLMs acquire sufficient task-specific semantics, given that such semantics are embedded in the fine-tuning data. During fine-tuning, the model continuously updates its internal representations through gradient descent, making it more sensitive to the semantic cues of the target task and achieving a more fine-grained understanding. Consequently, the fine-tuned smaller LLMs outperform the larger LLMs without fine-tuning. 

However, these improvements remain limited. For tasks such as \textit{Manner of Collision} extraction, the model only needs to capture the overall semantics of the narrative by identifying the primary collision pattern. Plugging categories are semantically well-separated in the embedding space, LLMs perform well without requiring fine-grained discrimination of textual details. Even if minor information is missed, the model can still predict the correct overall class. In contrast, \textit{Crash Type} requires accurate per-vehicle prediction among interwoven descriptions. To achieve this goal, the model has to carefully identify the target entity, accurately track target vehicles’ motion states, and resolve causal relations between events, while resisting misleading information from other vehicle descriptions. Here, even small errors such as confusing subjects and objects can lead to misclassification as shown in ~\ref{app:fail-case-2veh}. Although the self-attention mechanism enables the capture of long-range dependencies, it does not guarantee that all necessary semantics for answering the question are correctly identified. At the same time, the model is not fully resilient against irrelevant semantic details, making it prone to interference when vehicle descriptions are densely interwoven. As a result, LLMs are less effective at per-vehicle \textit{Crash Type} classification compared to per-crash \textit{Manner of Collision} classification. A promising solution is to incorporate attention guidance methods~\citep{wang-yu-2025-iquest, tianselective}, which can steer the model toward finer-grained semantic cues and enhance its ability to identify critical details.

\textbf{Enhancing crash data annotation}. Through the analysis of PLM-annotated data, we found that PLMs can successfully reclassify crashes originally labeled as \textit{Unknown} in the CISS dataset into specific categories (\ref{app:unknown2sepc}). They can also correct occasional labeling errors in the existing crash data (\ref{app:ex-1veh}), even though such errors are relatively rare. The high level of agreement across different models further supports the reliability of these classifications. 
Besides, we also studied if fine-tuned models introduce bias when classifying. In the two-vehicle setting, the comparison between CISS annotations and model predictions shows that fine-tuned models preserve the relational characteristics of crash data without introducing distributional bias. Overall, fine-tuned PLMs show strong potential to augment human annotation, correct errors, and improve the quality and reliability of crash datasets.

\textbf{Data quality, quantity, and robust model performance}. When varying the noise ratio in the training data, we find that all models remain robust under moderate label noise ($\leq$ 30$\%$). This is because, when noise is random and class-independent, the gradients from incorrect labels tend to cancel out in expectation, while the remaining 70\% of clean samples dominate learning and keep the decision boundary aligned with the correct direction. However, as the noise ratio increases further, LLM performance drops sharply due to conflicts between knowledge encoded in the prompt and the noisy training signals. 
In terms of training data scale, BERT and FastText perform better on very small datasets, whereas LLM is data-hungry. Because training involves additional prompts, its improvements are slower and it requires more data to converge. However, once the training set exceeds, the LLM exhibits an “aha” phase, with performance rising sharply and surpassing the other models. These results on relatively simple tasks highlight a trade-off among data quality, data quantity, and model performance. 

\section{Conclusion}
This work demonstrates that compact open-source PLMs, after fine-tuning with task-specific knowledge, can provide state-of-the-art performance for reasoning-intensive extraction from crash narratives, a long-standing challenge in traffic safety research where accurate analysis of crash is essential for crash prevention strategies and policy making. Our BERT and 3B model outperforms GPT-4o and LLaMA3-70B in identifying per-crash collision manner and per-vehicle \textit{Crash Type} in multi-vehicle scenarios, while requiring only limited training steps, modest resources, and delivering fast inference. Consistency analyses further show strong robustness across models and scenarios, enabling reliable per-vehicle \textit{Crash Type} labeling from complex narratives. Through the analysis of LLM-annotated data, we found that fine-tuned PLMs can correct labeling errors, address deficiencies caused by limited human knowledge, and preserve relational characteristics of the data without introducing bias. In addition, the approach mitigates privacy concerns associated with APIs and reduces deployment cost. Furthermore, experiments with varying levels of label noise demonstrate the robustness of PLMs, while scaling the amount of training data further highlights their potential to reduce human effort in data annotation.

All the above prove that automated annotation using PLMs is an alternative to labor-intensive manual coding. While our evaluation focuses on the U.S. CISS dataset and two extraction tasks, future work will explore broader safety information extraction (e.g., causal chains), uncertainty calibration, and aligning different annotation standards for comparative analysis. Overall, domain-adapted open-source PLMs offer a resource-efficient, privacy-preserving, and robust solution to improve traffic safety research through scalable crash narrative analysis. 
\section{Future work}

\textbf{Generalization capability.} Our study focused on extracting travel direction, \textit{Manner of Collision}, and \textit{Crash Type} from CISS crash narratives. Other attributes, such as impact points, roadway characteristics, and driver behavior, are also important for understanding collisions. Future work should extend our approach to these attributes and evaluate it on diverse datasets, including international crash records and social media reports.

\textbf{Reasoning-oriented LLMs.} Our experiments used BERT and instruction-tuned LLMs that directly output predictions. Although we used Chain-of-Thought prompting for LLMs, the models did not explicitly generate intermediate reasoning steps, nor did we adopt reasoning-specific models so as to reduce computation. Prior studies have shown that encouraging LLMs to articulate their reasoning process can improve prediction accuracy. In addition, during fine-tuning, it is possible to introduce guidance tokens that help the model learn to structure its reasoning. However, these approaches will also increase the computational costs.

\section*{Glossary}
For clarity, all field-specific terms used in this paper are summarized in Table~\ref{tab:glossary-traffic} and ~\ref{tab:glossary-nlp}, and all mathematical symbols are summarized in Table~\ref{tab:notation}.

\begin{table}[]
\footnotesize
\centering
\renewcommand{\arraystretch}{1.4} 
\caption{Traffic safety field-specific terms used in this paper.}  
\begin{tabular}{p{2.5cm} p{10cm}}

\hline
\textbf{Term } & \textbf{Description} \\
\hline

MANCOLL &
Indicates the manner in which vehicles in transport collided. \\

SUMMARY &
A basic text description of the crash scenario, and it may include special circumstances not captured in the
normal case coding. \\

CRASHCAT, CRASHCONF &
Variables Crash Category and Crash Configuration are used for categorizing the collisions of drivers involved in crashes. Each Category is further defined by a Crash Configuration.  \\

CRASHTYPE &
A numeric value used to classify the first harmful event in a crash and assigned by selecting the Crash Category and the Crash Configuration. \\
\hline
\end{tabular}
\label{tab:glossary-traffic}
\end{table}

\begin{table}[]
\footnotesize
\centering
\renewcommand{\arraystretch}{1.4} 
\caption{ Natural language processing field-specific terms used in this paper.}  
\begin{tabular}{p{4.4cm} p{8.1cm}}

\hline
\textbf{Term } & \textbf{Description} \\
\hline

Pre-trained language models &
Neural networks trained on vast amounts of text data, enabling them to learn general linguistic knowledge. \\

Chain-of-thought~\citep{wei2022chain} &
A prompt engineering technique that guides LLMs to break down complex tasks into intermediate steps. \\

Low-Rank Adaptation~\citep{hu2022lora} & Enable parameter-efficient fine-tuning of open-source LLMs, supporting safe and on-premise deployment.\\

\hline
\end{tabular}
\label{tab:glossary-nlp}
\end{table}

\begin{table}[htbp]
\footnotesize
\centering
\renewcommand{\arraystretch}{1.4} 
\caption{Summary of symbols and their descriptions used in this paper.}
\begin{tabular}{ll}
\hline
\textbf{Symbol} & \textbf{Description} \\
\hline
$X$ & Input hidden states, $X \in \mathbb{R}^{T \times d}$ ($T$: token length, $d$: hidden size) \\
$Q$ & Query matrix in self-attention, $Q = XW_Q$ \\
$K$ & Key matrix in self-attention, $K = XW_K$ \\
$V$ & Value matrix in self-attention, $V = XW_V$ \\
$W_Q, W_K, W_V$ & Projection matrices for query, key, and value \\
$A, B$ & Low-rank matrices in LoRA update ($A \in \mathbb{R}^{d \times r}$, $B \in \mathbb{R}^{r \times k}$) \\
$\Delta W$ & Low-rank update in LoRA: $\Delta W = \tfrac{\alpha}{r}AB$ \\
$W'$ & Adapted weight: $W' = W + \Delta W$ \\
$h_{\mathrm{CLS}}$ & Hidden representation of the [CLS] token from BERT encoder \\
$\mathcal{L}_{\mathrm{CE}}$ & Cross-entropy loss: $\mathcal{L}_{\mathrm{CE}} = - \log P(y|x)$ \\
$\mathrm{softmax}$ & Normalization function turning logits into probabilities \\

\hline
\end{tabular}
\label{tab:notation}
\end{table}

\section*{Funding sources}
Part of this research was supported by the Chalmers Transport Area of Advance project TREND.
\section*{Acknowledgements}
The authors would like to thank e-Commons (research infrastructure at Chalmers University of Technology) for their consultations, and extend special thanks to Nora Speicher and Mattia Carlino for valuable discussions related to running the experiments. We acknowledge the National Academic Infrastructure for Supercomputing in Sweden (NAISS), partially funded by the Swedish Research Council through grant agreement no. 2022-06725, for awarding this project access to the Alvis  cluster for computations and data handling.

\appendix
\section{Prompt in use}
\label{prompt-all}
\subsection{Prompt for \texttt{MANCOLL} classification}
\label{mancoll-prompt}
Task Introduction:

\vspace{6pt}
\begin{minipage}{\linewidth}
\small\itshape
 You are a helpful assistant that classifies vehicle collisions into one
of the following categories based on the description provided. Please
choose the most accurate collision type based on the definitions
and clarifications below:
\end{minipage}

\vspace{6pt}
Category Definitions (Expert Knowledge):
\vspace{6pt}

\begin{minipage}{\linewidth}
\small\itshape

{{
  0: "Not Collision with Vehicle in Transport - The vehicle did not
  collide with another vehicle in motion.",
  1: "Rear-End - The front of one vehicle strikes the rear of another
  vehicle traveling in the same direction.",
  2: "Head-On - The front ends of two vehicles traveling in opposite
  directions collide.",
  4: "Angle - The front of one vehicle strikes the side of another 
  at an angle (usually near intersections or crossing paths).",
  5: "Sideswipe, Same Direction - Both vehicles are moving in the same
  direction and **their sides make contact**",
  6: "Sideswipe, Opposite Direction - Both vehicles are moving in
  **opposite directions** and their **sides make contact**",
  9: "Unknown - The manner of collision cannot be determined."
}}
\end{minipage}

\vspace{6pt}
Clarification Rules (Special Prompting Instructions)
\vspace{6pt}

\begin{minipage}{\linewidth}
\small\itshape

If the collision happens at or near an intersection, classify as 4. 
If it does not occur near an intersection:
and both vehicles are traveling in the same direction, classify as 5.
and vehicles are traveling in opposite directions, classify as 6.
If the collision involves only one vehicle and a non-vehicle object
(e.g., animal, fence, tree), classify it as 0.
If no collision is described or it is unclear whether any impact occurred,
classify as 9.
If multiple collisions occur (e.g., chain reaction), classify based on
the **first** collision described in the summary.
\end{minipage}

Input original Extracted summary:
\begin{verbatim}
        \"\"\"{summary}\"\"\"
\end{verbatim}

\vspace{6pt}
Output Instruction (Task Constraint)
\vspace{6pt}

\begin{minipage}{\linewidth}
\small\itshape
Only respond with a single number from the list above.
Do not add any explanation.
    
\end{minipage}

\subsection{Prompt for \texttt{CRASHTYPE} classification}
\label{mancoll-prompt}
Task Introduction:

\vspace{6pt}
\begin{minipage}{\linewidth}
\small\itshape
 You are a crash analysis assistant. Your task is to assign a \textit{Crash Type} ID to a specific vehicle involved in a traffic crash, based on the structured context and detailed textual description below.

\end{minipage}

\vspace{6pt}
Category Definitions (Expert Knowledge):
\vspace{6pt}

\begin{minipage}{\linewidth}
\small\itshape
 Use the following \textit{Crash Type} definitions for classification:
 
\begin{verbatim}
    {crash_type_options} # This is decided by CRASHCONF
\end{verbatim}

\end{minipage}

\vspace{6pt}
Clarification Rules (Special Prompting Instructions)
\vspace{6pt}

\begin{minipage}{\linewidth}
\small\itshape
Crash classification must be based on the following inputs:
 \begin{verbatim}
        - Vehicle index: {vehicle_index}
        - Vehicle Description:
        \"\"\"{vehicle_summary}\"\"\"
\end{verbatim}

{Instructions:

- Carefully identify and focus on vehicle index in the text.

- Consider not only this vehicle's motion and behavior but also how it interacted with other vehicles (e.g., which vehicle was backing, struck another, etc.).

- Use both the structured crash context and relevant textual evidence to determine the most appropriate crash type ID.
}
\end{minipage}

\vspace{6pt}
Output Instruction (Task Constraint)
\vspace{6pt}

\begin{minipage}{\linewidth}
\small\itshape
Respond with only one number or letter corresponding to the correct \textit{Crash Type} from the options above. Do not include any explanation or extra text.
    
\end{minipage}

\section{Case study}
\subsection{Correct Prediction under Wrong Label}
\label{app:ex-1veh}
Here is an example where the fine-tuned PLM provides a more accurate classification compared to the ground-truth. In this case, the narrative text shows no clear indication of vehicle loss of control. While the ground-truth labeled it as a loss-of-control departure (7), the fine-tuned model identified it as a controlled left roadside departure (6). According to the coding guidelines in CISS, the first departure event should be considered and “loss of control” is defined as situations where the vehicle spins off due to surface conditions, oversteer phenomena, or mechanical malfunctions. Here, the vehicle first departed to the left roadside without cause explanation. The subsequent loss of control occurred as a consequence of this departure, not as its cause. Furthermore, in cases of doubt, the guideline explicitly advises using code “06” (Left Roadside Departure, Drive Off Road). In this way, to classify this case correctly, the model must capture crash causality, distinguish first events from outcomes, and apply the definition of “loss of control” with coding rules. 

\vspace{6pt}
\begin{minipage}{\linewidth}
\small\itshape
\textbf{SUMMARY}: V\#1 was traveling east, negotiating a curve left, in lane 1 of a 2 lane, two way roadway. \textbf{V\#1 departed the roadway to the left}, entered a drainage ditch area, and contacted an embankment with its front plane. The impact caused the vehicle to rotate in a counter clockwise direction where its front plane contacted the embankment a second time. The rear of V\#1 then elevated vertically into the air and continued to rotate counter clockwise as it re-entered the roadway. The vehicle then traveled backwards, departed the roadway to the left a second time, and began to rotate in a clockwise direction. The vehicle then rolled over one quarter turn onto its right side plane against an embankment, and came to final rest approximately 9 meters from the initial contact point, facing west.

\textbf{Ground Truth}: 7 -- Left roadside departure because of lost traction or control. (Not True)\\
\textbf{Mistral-7B fine-tuned with 1251 steps (LLM)}: 6 -- Left roadside departure under a controlled situation. (True)\\
\end{minipage}

To illustrate the distinction between controlled departures and loss-of-control scenarios, we present the following example, which clearly represents a loss-of-control crash. In this case, Vehicle \#1 departed its lane to the left and began rotating clockwise, providing clear evidence of a loss-of-control condition. Both the ground-truth label and the fine-tuned model correctly classified this as “7 — Left roadside departure because of lost traction or control”. 

\vspace{6pt}
\begin{minipage}{\linewidth}
\small\itshape
\textbf{SUMMARY}: V\#1 was traveling in a northern direction, on a two lane, non divided, bidirectional, dirt roadway with a downward grade, and a curve to the right. \textbf{V\#1 left its lane to the left, where it began to rotate clockwise}. V\#1 left the roadway to the left, where it overturned 6 quarter turns coming to final rest on its roof in the south bound lane of travel.

\textbf{Ground Truth}: 7 -- Left roadside departure because of lost traction or control. (True)\\
\textbf{Mistral-7B fine-tuned with 1251 steps (LLM)}: 7 -- Left roadside departure because of lost traction or control. (True)\\ 
\end{minipage}

\subsection{From \textit{Unknown} to Specific}
\label{app:unknown2sepc}

Here is an example that illustrates how fine-tuned PLMs can successfully resolve instances labeled as \textit{Unknown} in the CISS dataset. Despite the ground truth being marked as 9 (\textit{Unknown}), the model correctly identified the collision type as a Angle crash, demonstrating fine-tuned LLMs can identify and correct errors in the existing labels of crash data.

\vspace{6pt}
\begin{minipage}{\linewidth}
\small\itshape
\textbf{SUMMARY}: V2 was stopped southbound in a three lane intersection in lane three, awaiting to turn left eastbound. V1 was traveling westbound on a three lane road in lane two approaching the same intersection. V1 entered the south bound travel lanes of the intersecting roadway, where the front of V1 impacted the left side of V2.

\textbf{Ground Truth}: 9 -- \textit{Unknown} \\
\textbf{Mistral-7B fine-tuned with 1251 steps (LLM)}: 4 -- Angle. (True)
\end{minipage}

\subsection{Crash type classification in multi-vehicle scenario}
\label{app:success-case-5veh}

Here is an example that illustrates why fine-tuned LLMs perform better in scenarios involving more than two vehicles. The reason is that many vehicles beyond the second one are categorized into the class 98 (Third or subsequent vehicles involved in a crash). This aggregation reduces the classification granularity for such vehicles, effectively lowering the difficulty of prediction. As a result, when multiple vehicles are present, the model can rely on this broader category assignment, which improves overall performance compared to the more fine-grained distinctions required in two-vehicle crashes.

\vspace{6pt}
\begin{minipage}{\linewidth}
\small\itshape
\textbf{SUMMARY}: V1 was traveling north in the \#1 lane of a 5 lane divided roadway. V2 was stopped facing west in the \#2 lane of a 5 lane undivided roadway. V3 was stopped facing west in the \#1 lane of a 5 lane undivided roadway. V5 was stopped facing south in the \#5 lane of a 5 lane divided roadway. V1 traveled through the intersection. V2 and V3 started to accelerate and entered the intersection. The front of V1 impacted the left of V2. V2 rotated clockwise and the front of V2 impacted the left side of V3. V1 continued through the intersection where the front of V1 impacted the center median curb. V1 continued forward where the front of V1 impacted the left side of V5.\\

\textbf{Ground Truth}:\\
Vehicle 1:  88 -- Straight paths striking from the left\\
Vehicle 2:  89 -- Straight paths struck on the left\\
Vehicle 3: 98 --  Third or subsequent vehicles involved in a crash\\
Vehicle 4: 98 --  Third or subsequent vehicles involved in a crash\\
Vehicle 5: 98 --  Third or subsequent vehicles involved in a crash\\

\textbf{LLaMA3-8B fine-tuned with 2163 steps (LLM)}:\\
Vehicle 1:  88 -- Straight paths striking from the left (True)\\
Vehicle 2:  89 -- Straight paths struck on the left (True)\\
Vehicle 3: 98 --  Third or subsequent vehicles involved in a crash (True)\\
Vehicle 4: 98 --  Third or subsequent vehicles involved in a crash (True)\\
Vehicle 5: 98 --  Third or subsequent vehicles involved in a crash (True)\\
\end{minipage}

\subsection{Crash type classification in two-vehicle scenario}
\label{app:fail-case-2veh}

In this example, the fine-tuned LLaMA3-8B model fails to correctly distinguish the \textit{Crash Types} of V1 and V2. The primary difficulty arises from the narrative, where the descriptions of V1 and V2 are closely intertwined, making it hard for the model to disentangle their respective roles. Moreover, the fact that both vehicles rotated and came to rest in similar orientations further complicates classification, as these post-crash states provide limited discriminative cues. The only decisive difference is that V1 rear-ended V2, while V2 was rear-ended by V1. This subtle asymmetry requires precise tracking of subject–object relations in the narrative. The model’s prediction error suggests that it could not fully capture all semantics encoded in the prompt, highlighting a key limitation of LLMs in fine-grained, semantic-sensitive classification.

\vspace{6pt}
\begin{minipage}{\linewidth}
\small\itshape
\textbf{SUMMARY}: V1 was negotiating a right curve traveling east in the first of four lanes on a divided highway. V2 was traveling east in the same lane ahead of V1. The front of V1 contacted the back of V2. V1 rotated counterclockwise and came to rest on the roadway facing northeast. V2 rotated counterclockwise and came to rest on the roadway facing northeast.\\

\textbf{Ground Truth}:\\
Vehicle 1:  24 -- Rear-end: slower (a vehicle that impacts another vehicle from the rear
when the struck vehicle was going slower than the striking vehicle.)\\
Vehicle 2:  25 -- Rear-end: slower, going straight (a rear-ended vehicle that was going slower than the other vehicle while proceeding straight ahead.)\\

\textbf{LLaMA3-8B fine-tuned with 2163 steps (LLM)}:\\
Vehicle 1:  24 -- Rear-end: slower (True)\\
Vehicle 2:  24 -- Rear-end: slower (False)\\

\end{minipage}


\bibliographystyle{elsarticle-num-names} 
\bibliography{cas-refs}






\end{document}